\documentclass[10pt,journal,compsoc]{IEEEtran}

\usepackage{graphicx}
\usepackage{amsfonts}
\usepackage{amsmath}
\usepackage{amssymb}
\usepackage{diagbox}
\usepackage{multirow}
\usepackage{graphics}
\usepackage{epstopdf}
\usepackage{tabularx}
\usepackage{epsfig}
\usepackage{color,subfigure}
\usepackage{times}
\usepackage{mathrsfs}
\usepackage{bm}
\usepackage{algorithm, algorithmic}
\usepackage{mathdots}
\usepackage[american]{babel}
\usepackage{microtype} 
\usepackage{array}
\usepackage{setspace}
\usepackage{threeparttable}
\usepackage{url}
\usepackage{color}
\usepackage[colorlinks,linkcolor=red]{hyperref}

\ifCLASSOPTIONcompsoc

  \usepackage[nocompress]{cite}
\else
  \usepackage{cite}
\fi

\ifCLASSINFOpdf
\else
\fi

\begin{document}

\title{Adversarial Sticker: A  Stealthy Attack Method \\in {the} Physical World}

\author{Xingxing Wei, 
        Ying Guo,
        and~Jie Yu
\IEEEcompsocitemizethanks{
\IEEEcompsocthanksitem Xingxing Wei, Ying Guo and Jie Yu were with the Institute of Artificial Intelligence, Beihang University, No.37, Xueyuan Road, Haidian District, Beijing, 100191, P.R. China.
(E-mail: \{xxwei, yingguo, jieyu\}@buaa.edu.cn)
\IEEEcompsocthanksitem Xingxing Wei is the corresponding author}
}


\IEEEtitleabstractindextext{%
\begin{abstract}
  To  assess the vulnerability of deep learning in the physical world, recent works introduce adversarial  {patches} and apply  {them} on different  tasks. In this paper, we propose another kind of adversarial patch:  {the} Meaningful  Adversarial Sticker, a physically feasible and  stealthy attack method by using real stickers existing in our life.  {Unlike} the previous adversarial patches by designing perturbations, our method manipulates the sticker's pasting {position and rotation} angle on the objects to perform physical attacks. Because the  position and rotation angle are less affected by the printing loss and color distortion, adversarial stickers can keep good attacking performance in the physical world.  Besides, to make adversarial stickers more practical in real {scenes},  we conduct attacks in the black-box setting with  {the} limited information rather than the white-box setting with all the details of threat models. To effectively solve  {for} the sticker's  parameters, we design  {the} Region based Heuristic Differential  {Evolution} Algorithm, which utilizes the new-found regional aggregation of effective solutions and the adaptive adjustment strategy of  {the} evaluation criteria. Our method is comprehensively verified in the face recognition and then extended to the image retrieval and traffic sign recognition. Extensive experiments  show  the proposed method is effective and efficient in complex physical conditions and {has}  {a} good generalization for different tasks. 
   
\end{abstract}

}

\maketitle

\IEEEdisplaynontitleabstractindextext

\IEEEpeerreviewmaketitle

\IEEEraisesectionheading{\section{Introduction}\label{sec:introduction}}

\IEEEPARstart{W}{ith} the development of Deep Neural Networks (DNNs), DNNs based vision systems have shown  {the} excellent performance in different tasks \cite{guo2020deep,minaee2021image,baltruvsaitis2018multimodal}. However, DNNs are vulnerable to  adversarial examples \cite{szegedy2013intriguing, goodfellow2014explaining}. By adding a small malicious perturbation to the input example, the system can make a wrong identity judgement, resulting in serious consequences.

In real scenarios, however, DNNs based vision systems work by directly scanning objects. So attackers can only change the  {object's} appearance in the physical world to provide malicious inputs to the camera, which is more challenging and needs to tackle complex physical conditions such as lighting, distance, and posture changes.
To make adversarial examples available in the physical world, recent works introduce the adversarial patch \cite{brown2017adversarial}. They {do not} restrict the perturbations' magnitude, and generate adversarial perturbations in a fixed region. Experiments show  the patch can be placed on the objects, and causes the classifier to output a targeted class.
Up to now, researchers have applied adversarial patches on different tasks, such as face recognition \cite{dong2019efficient,sharif2016accessorize,komkov2019advhat,sharif2019general}, object  {detection} \cite{liu2018dpatch,zhao2020object}, pedestrian detection \cite{thys2019fooling,xu2020adversarial}, etc.


Despite the success of adversarial patches, they have two limitations. The \textbf{first one} is that the patch's perturbation pattern will face a complex transfer process from the digital domain to the physical world. Specifically, Expectation Over Transformation (EOT) \cite{athalye2018synthesizing}, Total Variation (TV) loss, and non-printability score (NPS) loss \cite{komkov2019advhat,sharif2019general} {are used} to ensure the attacking performance of real-world adversarial examples. EOT considers a set of transformations of objects (postures, distance changes, etc.) when generating adversarial perturbations.  {The} TV loss is designed to make the perturbations more smooth, and  {the} NPS loss is to deal with the difference between digital pixel values and the actual printed appearance. On the one hand, these operations lead to high computation costs, for example, EOT needs to exhaust different transformations. On the other hand, the perturbations' values will inevitably become distorted due to the limitation of printing devices despite the usage of  {the} TV and NPS losses. Last but not least, the current physical perturbations are meaningless and irregular,  {not natural enough in the appearance, and prone to arouse the human's suspicion}. These disadvantages motivate us to explore new forms of adversarial  {patches} to address the above issues. The \textbf{second one} is that most previous adversarial patches are based on the white-box attack setting \cite{brown2017adversarial,thys2019fooling,xu2020adversarial,komkov2019advhat,sharif2016accessorize}. It means that they require detailed structures and parameters of the targeted models. However, {the above} information usually cannot be easily obtained especially in the actual applications. Instead, some commercial online vision APIs (e.g. Face++ and Microsoft cloud services) usually return the predicted identities and scores for the uploaded images. By utilizing {this} limited information, exploring  {the} query-based black-box attack to construct  adversarial  {patches} is a more reasonable solution in most real scenarios than the previous white-box attacks. 


Considering the above two points, this paper aims to solve the following problem: 
\emph{under the black-box setting with  {some} limited information, how to utilize the existing material in our life to easily construct {a} stealthy adversarial patch which is robust to the complex physical changes.}

\begin{figure}[t]
\centering\includegraphics[width=0.49\textwidth]{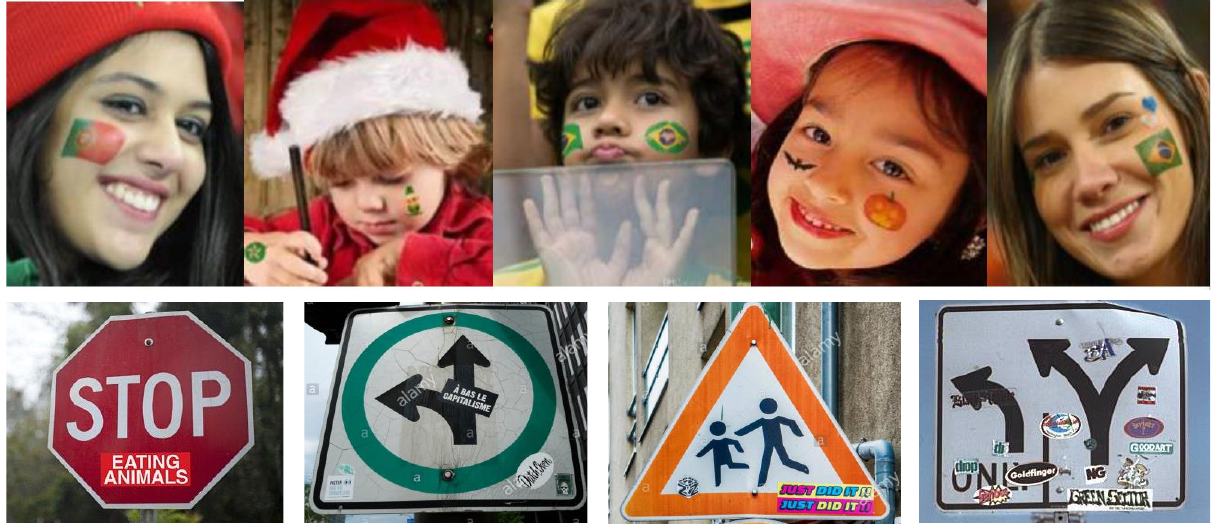}
\caption{Examples of {pasting} stickers on the face and traffic sign in our life, respectively.}
\label{fig:life}
\end{figure}

In this paper, we propose to construct a novel form of  adversarial patches called  {the} Meaningful Adversarial Sticker via  {a} query-based black-box attack. Instead of generating  {an} adversarial perturbation pattern like the traditional adversarial patch, we use the real meaningful stickers existing in our life and manipulate the stickers' pasting positions and rotation angles on the objects to perform the physical attacks. 
Compared with the performance due to perturbations, the attack performance caused by stickers' positions and rotation angles is easier to maintain when attacks are transferred from  {the} digital domain to the physical world (see Section \ref{sec:imp_phy}), and the loss functions mentioned before (EOT, TV and NPS losses,  {etc.}) are not necessary.
Furthermore, sticking colorful stickers on objects can be usually seen in our life (shown in Figure \ref{fig:life}), so this form of attacks  looks natural and  stealthy, and is not easy to arouse the human's suspicion. In addition, considering the real scenario where only  {the} limited information can be obtained, we design a query-based method to efficiently search for the available parameters. Thus the physical black-box attacks are achieved. 


Technically, to search for the appropriate attack parameters, we formalize the process into an optimization problem and solve it using an evolution method which follows the principle of ``survival of the fittest" in the iterative evolution process. Considering the query limit in the actual scenario, we design a new \textbf{Region based Heuristic Differential  {Evolution} (RHDE)} algorithm to improve the efficiency. We find that the stickers' locations of successful attacks show {a} regional aggregation. Based on this, RHDE combines  {the} inbreeding and random crossover to generate the offspring, and adjusts the evaluation criteria adaptively to better guide the search direction. We also design a sticker deformation calculation method to make the  {sticker's} shape fit the curvature changes of different positions on the human face realistically. The proposed method is first verified on  {the} face recognition, and then extended to  {the} image retrieval and traffic sign recognition tasks. The code is available at \url{https://github.com/jinyugy21/Adv-Stickers_RHDE}.

In summary, this paper has the following contributions:
\begin{itemize}
\setlength{\itemsep}{0pt}
\item We propose  {the} Meaningful Adversarial Sticker, a novel adversarial patch method  with the good practical applicability. We manipulate the fusing operation and parameters of real stickers on the objects instead of designing perturbation patterns like most of the existing works. Experiments show this manner has  {a} good transferability from  {the} digital domain to the physical world.
\item We specialize in black-box physical attacks on DNNs based vision systems with  {some} limited information, and further design a Region based Heuristic Differential  {Evolution} (RHDE)  algorithm to improve  {the} query efficiency. We find that the stickers' locations of successful attacks show  {a} regional aggregation. RHDE makes full use of this phenomenon and can adjust the evolution direction adaptively according to the state of the population.
\item We verify the proposed method on three tasks: face recognition, image retrieval, and traffic sign recognition. For the face recognition, experimental results in the physical environment show that it can naturally maintain attack effects under different physical conditions and at most 98.46\% of the video frames can be successfully attacked while continuously changing the face postures. For the image retrieval and traffic sign recognition, our method also {shows}  {a} good generalization. 
\end{itemize}

 \par The remainder of this paper is organized as follows. Section 2 briefly reviews the related work. We introduce the details of the proposed meaningful adversarial stickers against the face recognition task in Section 3.
 Section 4  {presents extensions to the image retrieval and traffic sign recognition tasks}. Section 5 and 6 show a series of experimental
 results. Finally, we conclude the whole paper in Section 7.

\section{Related Work}

In this section, we review the existing works of adversarial examples in the digital and physical world, respectively.
\subsection{Digital Attacks}
Box-constrained L-BFGS \cite{szegedy2013intriguing}, C\&W \cite{carlini2017towards}, Deepfool \cite{moosavi2016deepfool}, etc. carry out attacks via optimization mechanisms. The classical attack method FGSM \cite{goodfellow2014explaining} is a one-step approach based on the gradient information of DNNs.  {PGD \cite{madry2017towards} is a multi-step iterative method using projected gradient descent on the negative loss function to generate adversarial examples.} Instead of  {generating} noises on the whole image, \cite{karmon2018lavan} explores the case where the noise is allowed to be visible but confined to a small, localized patch of the image, without covering any of the main object(s) in the image. The above methods are  {conducted} in the white-box setting, where the attackers have access to the structures and weights of the threat models.

Black-box attacks do not require detailed parameters of models. For transfer-based methods, the adversarial examples generated for one model can be transferred to another model to achieve successful attacks \cite{dong2018boosting,liu2016delving,chen2020universal,wang2020hamiltonian,wu2018understanding}. For score-based methods, the probabilities predicted by target models are known and methods such as gradient estimation \cite{chen2017zoo} and random search \cite{andriushchenko2020square,guo2019simple} are often used in this setting. Besides, decision-based methods are suitable for more restrictive scenarios where only the final model decisions are known \cite{brendel2018decision,cheng2018query}. Dong \emph{et al.} \cite{dong2019efficient} conduct digital attacks on face recognition systems in this setting and model the local geometry of solving directions to improve  {the} efficiency. Attacks aiming to get a different class from the true label are called un-targeted attacks (or \textbf{dodging} in the face recognition),  while those targeting a specific class are called targeted attacks (or \textbf{impersonation} in the face recognition).
In our case, we conduct black-box attacks and focus on more practical physical attacks.


\subsection{Physical Attacks}
Physical attacks play an increasingly important role due to their great application values.  To make adversarial examples available in the physical world, many works have been proposed. Specifically, Kurakin \emph{et al.} \cite{kurakin2016adversarial} verify the feasibility of physical attacks by the fact that the perturbed images being captured by the camera still have attack effects. In \cite{athalye2018synthesizing}, the EOT algorithm makes adversarial patches robust to multiple physical transformations.

Adversarial patch \cite{brown2017adversarial} is one of the main forms of physical attacks, and has been successfully applied in many computer vision tasks, such as automatic driving, face recognition, object  {detection}.  We give the detailed descriptions as follows:

For face recognition cases, the initial attack is in the form of 2D-printed face photos or 3D facial masks \cite{Hernandez-Ortega2019}. Later, some researchers generate eyeglass frames with perturbations attached to fool the face recognition systems \cite{sharif2016accessorize,sharif2019general}. 
Zhou \emph{et al.} \cite{zhou2018invisible} carry out attacks by a cap mounting LED device which illuminates the face using infrared perturbations. 
Adv-Hat \cite{komkov2019advhat} achieves attacks by sticking rectangular stickers with adversarial perturbations to the hat. Adversarial light projection attacks \cite{nguyen2020adversarial} project transformation-invariant adversarial patterns onto people's faces. Some other methods \cite{kaziakhmedov2019real,pautov2019adversarial} paste black and white patches with an attacking effect onto the face or  {the} wearable accessory.

For automatic driving  {tasks}, Eykholt \emph{et al.} \cite{eykholt2018robust} use Robust  Physical  Perturbations to generate  {the} adversarial graffiti which is robust under physical conditions. \cite{zolfi2021translucent}  proposes the physical translucent patch, placed on the camera lens, which results in the failure to detect the stop sign while correctly identifying the other objects. The authors in \cite{duan2020adversarial}  craft and camouflage physical-world adversarial examples into natural styles that appear legitimate to human observers to construct adversarial traffic  {signs}.  Several relevant studies considering the safety of the autonomous driving can also be found in \cite{eykholt2017note,sitawarin2018darts}.

For object  {detection tasks},  \cite{song2018physical}  extends physical attacks to more challenging object detection models and proposes {the ``Disappearance} Attack". The work in \cite{thys2019fooling} uses the adversarial patch to hide a person from a person detector.   Adversarial T-shirts \cite{xu2020adversarial} can apply the deformable adversarial patch on the T-shirts to fool the person detector. Wu\ \emph{et al.} \cite{wu2020making} present a detailed study of physical world attacks using printed posters and wearable
clothes, and quantify their performances  with different metrics.

In our method, we propose another kind of adversarial patch, which uses the real stickers  {that actually exist in our life}. The proposed adversarial stickers conduct black-box attacks by changing the real stickers' pasting parameters instead of the  {patches'} perturbations and thus do not need to be generated or printed, which is physically feasible and stealthy. 

\begin{figure}[t]
\centering\includegraphics[width=0.4\textwidth]{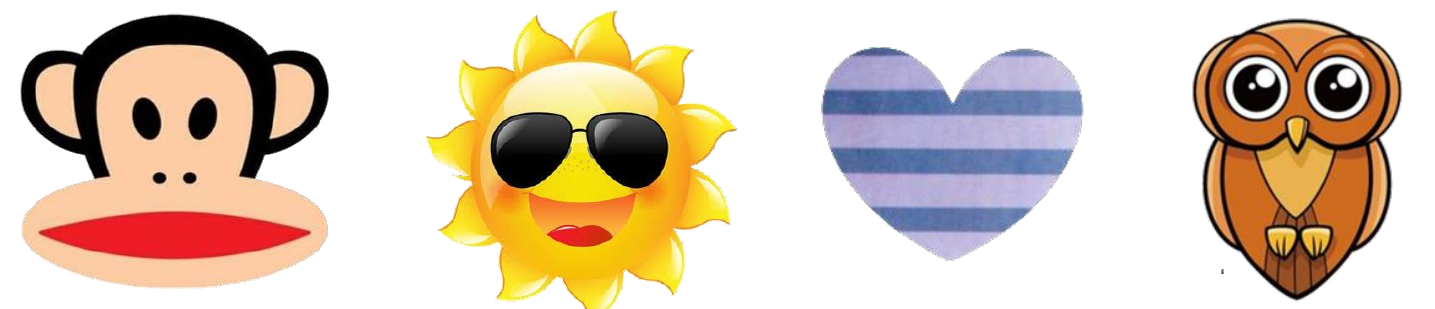}
\caption{Some examples of stickers used in our experiments.}
\label{fig:life1}
\end{figure}

\section{Methodology}\label{sec:method}
In this section, we choose  {the} face recognition  as the target task, and first introduce the regional aggregation of effective parameters, then detail the proposed Region based Differential Evolution algorithm, and finally introduce the calculation method of  {the} sticker deformation in the process of parameter solving.

\subsection{Problem formulation}
In the face recognition task, given {a} clean face image $\bm{x}$, the goal of the adversarial attack is to make the face recognition model predict a wrong identity of the perturbed face image $\bm{x}^{adv}$. Formally, the perturbed face with the adversarial patch can be formulated as Eq. (\ref{eq:xadv}), where $\odot$ is the Hadamard product and $\bm{\tilde x}$ is the adversarial  perturbations. $\mathcal{A}$ is a mask matrix to constrain the shape and pasting position of the patch, where the value of the pasting area is 1.
\begin{equation}\label{eq:xadv}
\bm{x}^{a d v}=(1-\mathcal{A}) \odot \bm{x}+\mathcal{A} \odot \bm{\tilde x}
\end{equation}
The previous methods mainly optimize $\bm{\tilde x}$ with  {a} pre-fixed $\mathcal{A}$. In contrast, our method does not generate $\bm{\tilde x}$,  {but chooses} the existing sticker in our life as $\bm{\tilde x}$, which is shown in Figure \ref{fig:life1}.  When the sticker is chosen, the shape of $\mathcal{A}$ is fixed. In the following section, we  {will show} how to obtain the optimal pasting position of $\mathcal{A}$ in the face to perform adversarial attacks.

\begin{figure}[t]
\centering\includegraphics[width=0.49\textwidth]{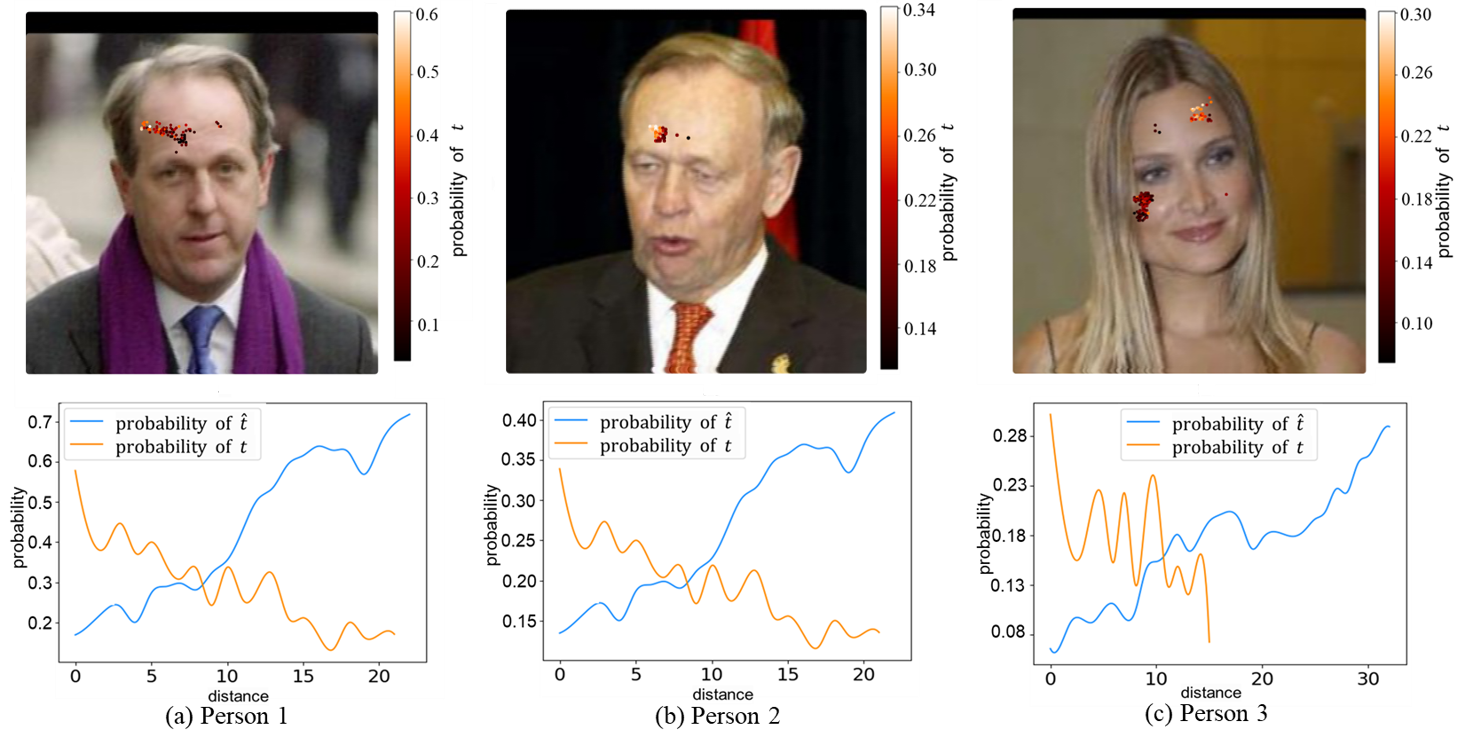}
\caption{ {Examples reflecting the regional aggregation of positions for successful attacks. The top row shows the distribution of these positions, and the bottom row shows {the probabilities of the ground-truth label $\hat{t}$ and predicted wrong label $t$} versus the distance to the center.}}
\label{fig:region}
\end{figure}

\subsection{Regional aggregation}
Before introducing the detailed method, we first explore the influence of pasting positions for  {the} face recognition task. 
In such scenes, liveness detection, which mainly relies on motions (e.g. blinking, mouth opening), depth or texture features of the face \cite{ming2020asurvey}, is often used to confirm the real physiological characteristics of the object and resists attacks such as photos, masks, and screen re-shoots.
To ensure  {a} natural look and not interfere with the liveness detection, pasting positions of stickers cannot cover the facial features. Thus, a face mask matrix $M^{F} \in R^{n \times m}$ which contains ones in valid regions (e.g. cheek and forehead), and zeros in invalid regions (e.g. eyes and mouth) is used to constrain the candidate pasting areas of stickers.

{We randomly select 1000 images from LFW and CelebA datasets respectively, and use the first sticker in Figure 2, then traverse every valid pasting position by  {an} exhaustive method to query the FaceNet \cite{schroff2015facenet} model.
It is found that the positions achieving successful attacks are not discretely distributed, but show  {an} aggregation phenomenon. Among all the faces that can be successfully attacked on the two datasets, the faces with clustered positions account for 96.5\% and 97.8\%, respectively, of which 58.7\% and 61.2\% are clustered in only one area (like Figure \ref{fig:region} (a), (b)), and 37.8\% and 36.6\% have multiple clustered areas (like Figure \ref{fig:region} (c)).}
We also analyze the probability variations of ground-truth labels $\hat{t}$ and predicted wrong labels $t$ after attacks when we take the point with the highest value of $t$ as the center $o^{*}$ and randomly choose one direction to spread outward.
In the small area around $o^{*}$, the probability of $t$ decreases with the increase of distance to $o^{*}$, while  {the trend for} the probability of $\hat{t}$ is the opposite.
Figure \ref{fig:region} shows several examples.

This phenomenon inspires us to search for the optimal position with a heuristic manner, i.e., continuing to search for the next position around the  {previously} available position. In this way, we can improve the efficiency of searching attack parameters. Based on this idea, a region based differential evolution algorithm is designed, which is shown in Section \ref{sec:rhde}.

\subsection{Region based  Heuristic Differential Evolution}\label{sec:rhde}
Let $f(\cdot)$ denote the face recognition model and $f(\bm{x},t)$ denote the probability that the model predicts a face image $\bm{x}$ as label $t$. $\bm{\theta}=\left(\theta_{1}, \ldots, \theta_{i}, \ldots, \theta_{d}\right)$ is the set of attack parameters (including pasting position, rotation angle, etc.). Given the ground-truth label $\hat{t}$ for $\bm{x}$ and a real sticker image $s$, the goal of a \textbf{dodging} (un-targeted) attack is to find the optimal attack parameters $\bm{\theta}^{*}$ to make the probability corresponding to $\hat{t}$ as small as possible, so that a person different from $\hat{t}$ is regarded as  {the} top-1 identity. So the objective function of  {the} dodging attack can be formalized as:
\begin{equation}\label{eq:untarget}
\min _{\bm{\theta}} \ \ \mathcal{L}_{\emph{dodging}}(\bm{\theta})=f(g(\bm{x} ; s, \bm{\theta}), \hat{t})
\end{equation}
where $g(\bm{x} ; s, \bm{\theta})$ represents the generated new face image after transforming sticker $s$ according to $\bm{\theta}$ and combining the obtained sticker with the face. Details of sticker transformation are shown in Section \ref{sec:transform}.

For  {the} \textbf{impersonation} (targeted) attack, given a target identity $t^{*}$, the objective function is defined as follows:
\begin{equation}\label{eq:target}
\min _{\bm{\theta}} \ \ \mathcal{L}_{\emph{impersonation}}(\bm{\theta})=1-f\left(g(\bm{x} ; s, \bm{\theta}), t^{*}\right)
\end{equation}

Since we have no access to the specific parameters of $f(\cdot)$, we carry out score-based black-box attacks by querying the model to obtain predicted labels and probabilities. Although gradient estimation \cite{chen2017zoo} can solve the optimization problem along the gradient descent direction, in our case, the ranges of position parameters are discontinuous due to the invalid positions. Accordingly, the objective functions are discontinuous and their smoothness  {with respect to} the parameters is also unknown, so it is not suitable to use  {the} gradient-based method to optimize Eq. (\ref{eq:untarget}) and Eq. (\ref{eq:target}). Therefore, we use an evolutionary method, starting from a group of randomly generated solutions in the search space and using  {the} crossover and mutation to generate  {the} offspring, making the fittest survive according to the evaluation criteria, and finally find the appropriate solution in the iterative evolution process.

However, using traditional evolutionary algorithms directly is not efficient enough because the characteristics of face recognition scenes are not fully considered. In this paper, we propose a novel \textbf{Region based Heuristic Differential  {Evolution} (RHDE) } algorithm  to accelerate the search for solutions. We design a new strategy for the offspring's generation, which utilizes  {the} regional aggregation of positions with attacking effectiveness. To better guide the search direction, we also use an adaptive evaluation criteria adjustment method to adjust the attack target in time according to the current state of the solutions. Taking  {the} dodging attack for example, the overall RHDE algorithm is outlined in Algorithm \ref{alg:1}. Details are shown in the following.

\subsubsection{Attack setting}
In the evolutionary approach, a population represents a set of multiple solution vectors and each individual in the population represents a solution vector. Given the population size $P$ and the number of attack parameters to be solved $d$, the $k$-th generation population $\bm{X}(k)$ is represented as: 
\begin{equation}\label{eq:population}
\bm{X}(k):=\left\{\bm{X}_{i}(k) |  \theta_{j}^{L} \leq \bm{X}_{i j}(k) \leq \theta_{j}^{U}, 1 \!\leq\! i \!\leq \!P, 1 \!\leq\! j\! \leq\! d\right\}
\end{equation}
where $\bm{X}_{ij}(k)$ is the $j$-th parameter value of the $i$-th individual in the $k$-th population. $\left(\theta_{j}^{L}, \theta_{j}^{U}\right)$ is the change range of the $j$-th parameter. Specifically, each individual in the population represents a tuple containing the pasting position, rotation angle, etc.
In each iteration, a new offspring is formed through  {the} crossover and mutation between individuals in the parent population. According to the evaluation criteria, better individuals are selected from the offspring and the parent to {create} the next generation (new solutions).

In Algorithm \ref{alg:1}, we first randomly initialize the population $\bm{X}(0)$ on the premise of ensuring that the parameters of each individual are within the corresponding value range (Step 1). Then we generate candidate populations $\bm{C}(k)$ in  {an} iterative evolution process (Step 7). Based on the evaluation criterion $\mathcal{J}(\bm{\theta})$, better individuals between $\bm{C}(k)$ and $\bm{X}(k)$ are chosen to form the next generation $\bm{X}(k\!+\!1)$. The process stops when the attack using the optimal individual in the current population as the attack parameters is successful (Step 4) or when the maximum number of iterations $T$ is reached. The generation strategy of $\bm{C}(k)$ and the establishment of $\mathcal{J}(\bm{\theta})$ are detailed in Section \ref{sec:offspring} and Section \ref{sec:criteria}.

\begin{algorithm}[t]
	\renewcommand{\algorithmicrequire}{\textbf{Input:}}
	\renewcommand{\algorithmicensure}{\textbf{Output:}}
	\caption{\small Region based Heuristic Differential  {Evolution} Algorithm}
	\label{alg:1}
	\begin{algorithmic}[1]
		\REQUIRE {Network $f(\cdot)$, face image $\bm{x}$ and label $\hat{t}$, the attack objective function $\mathcal{L}(\bm{\theta})$, the number of parameters $d$, value range {\footnotesize$\left(\theta^{L}, \theta^{U}\right)$}, population size $P$, maximum number of iterations {\footnotesize$T$}, hyperparameter $l$, $r$, $\alpha$, $\rho$, $\delta$}
		\ENSURE $\bm{\theta}^{*}$
		\STATE {\small Initialize $\bm{X}(0)$ randomly in $\left[\theta_{i}^{L}, \theta_{i}^{U}\right](1 \!\leq\! i \!\leq\! d)$, \\
		$\mathcal{J}(\theta)$ = $\mathcal{L}(\theta)$, $flag$ = 0, $stop$ = $T$};
		\FOR{$k$ = 0  to $T\!-\!1$}
    		\STATE {\small Sort $\bm{X}(k)$ in ascending order according to $\mathcal{J}(\theta)$;}
    		\IF{\small$\bm{X}_{0}(k)$ makes the attack successful}
        		\STATE{\small$stop$ = $k$; \ break;}
        	\ENDIF
    		\STATE Generate candidate population $\bm{C}(k)$\\
    		{\quad  {if} \textls[1]{\small$i \in[1, \mu \!*\! P]$} \ \small$\bm{C}_{i}(k)\!\leftarrow\!$ according to Eq. (\ref{eq:inbreeding}) \\
    		 \quad  {if} \textls[1]{\small $i \in[\mu \!*\! P\!+\!1, P]$} \ $\bm{C}_{i}(k)\!\leftarrow\!$ according to Eq. (\ref{eq:cross})}
            \IF{$\left(t_{1}\left(C_{\gamma^{*}}(k)\right)==\hat{t}\right.$ and $flag \left.==0\right)$}
                \STATE$ {bound}\!\leftarrow\!$ according to Eq. (\ref{eq:difference})
                \IF{$bound \leq \delta$}
                    \STATE {$flag\!=\!1, \bm{\tau}\!=\!t_{2}$}; {\small Update  $\mathcal{J}(\theta)$ according to Eq. (\ref{eq:update_j})}
                \ENDIF
            \ENDIF
    		\STATE {\small$\bm{X}_{i}(k+1) \leftarrow$ {\small the better one between} $\bm{X}_{i}(k)$ and $\bm{C}_{i}(k)$}
		\ENDFOR
		\STATE {\small Sort $\bm{X}(stop)$ in ascending order according to $\mathcal{J}(\theta)$;}
		\STATE {\small\textbf{return} $\bm{X}_{0}(stop)$ }
	\end{algorithmic}  
\end{algorithm}

\subsubsection{Strategies for the  {offspring's} generation}\label{sec:offspring}
In our proposed algorithm, we use \emph{crossover} between random individuals and \emph{inbreeding} of superior individuals to generate candidate populations $\bm{C}(k)$. The first method follows the traditional evolutionary algorithm, and can be formalized as follows:
\begin{equation}\label{eq:cross}
\bm{C}_{i}(k)=\operatorname{clip}\left(\bm{X}_{\gamma^{*}}(k)+\alpha\left(\bm{X}_{\gamma_{1}}(k)-\bm{X}_{\gamma_{2}}(k)\right)\right)
\end{equation}
where $\bm{C}_{i}(k)$ is the $i$-th individual in the $k$-th candidate population. $\gamma_{1},\gamma_{2}$ are random numbers. $\gamma^{*}$ denotes the index number of the best individual in $\bm{X}(k)$ and {\small$\gamma^{*} \neq \gamma_{1} \neq \gamma_{2}$}. $\alpha$ is  {a} scale factor and $\operatorname{clip}(\cdot)$ is a clipping operation to keep individuals within the range described in Eq. (\ref{eq:population}).

\begin{figure*}[th]
\centering\includegraphics[width=0.98\textwidth]{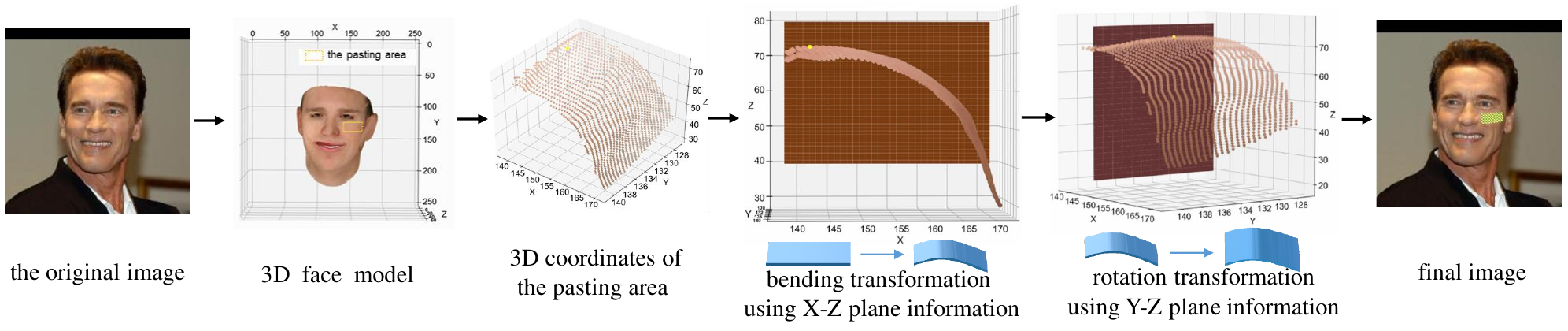}
\caption{The process of bending and rotating the sticker (the yellow dot indicates the highest point of the pasting area).}
\label{fig:procedure}
\end{figure*}

Because the solutions with adversarial effects tend to cluster in a certain region in the parameter space, we propose an \emph{inbreeding} method, which finds solutions in the regions near the superior solutions of each generation to speed up the solving process. Specifically, the superior individuals in the current population are selected first (by a ratio of $\mu$). $\bm{\phi}\langle\bm{X}_{i}(k), j, l\rangle$ is defined as an operation, which takes the position in individual $\bm{X}_{i}(k)$ as the center, takes out the position parameter at the step size $l$ in the $j$-th direction around the center, and {forms} a new individual together with the rest of the parameters in $\bm{X}_{i}(k)$. $\bm{\phi}$ is applied in {$r$} directions around the superior individuals (i.e. {\small$1 \leq j \leq r$}), and the individuals that  {minimize the loss function} (most satisfied with the objective function) around each superior individual are 
\noindent{selected as the offspring. The formula is defined as follows:}
\begin{equation}\label{eq:inbreeding}
\bm{C}_{i}(k)=\bm{\phi}\left\langle \bm{X}_{i}(k), {\arg\min }_{j} \mathcal{L}\left(\bm{\phi}\left\langle \bm{X}_{i}(k), j, l\right\rangle\right), l\right\rangle
\end{equation}
Combining these two methods, we take advantage of the regional aggregation of effective solutions  {by the inbreeding on the one hand, and generate more diverse solutions by a random method on the other hand, avoiding the local optimum that the inbreeding may fall into.}

\subsubsection{Adaptive adjustment of  {the} evaluation criteria}\label{sec:criteria}
{The solving process can be regarded as an exploration-exploitation process.}
For the evaluation criterion $\mathcal{J}(\bm{\theta})$ (the smaller the better), which evaluates the {unfitness} of individuals in a population, it is generally equal to the value of the objective function $\mathcal{L}(\bm{\theta})$ (Step 1).
For the dodging attack, we design an adaptive strategy to adjust the evaluation criterion (Step 8-13).
{In the initial stage, there is no selected target identity, so we let $\mathcal{J}(\bm{\theta})$ follow $\mathcal{L}(\bm{\theta})$ to reduce the predicted probability of  {the} ground-truth label $\hat{t}$, so as to explore as many solutions as possible in the parameter space.}
When the population evolves to a good state (i.e. the difference between the predicted probability of top-1 class $t_{1}$ and top-2 class $t_{2}$ corresponding to the optimal individual $\bm{C}_{\gamma^{*}}(k)$ is less than the threshold value $\delta$),
{we start to exploit the information of top-2 class $t_{2}$ and transform the solving direction to improving the predicted probability of $t_{2}$.}
 {After abbreviating} $f(g(\bm{x} ; s, \bm{\theta}), {t})$ to $f_{t}(\bm{\theta})$, the indicator to determine whether the population has reached a good state is formulated as:
\begin{equation}\label{eq:difference}
\mathit{bound}=f_{t_{1}}\left(\bm{C}_{\gamma^{*}}(k)\right)-f_{t_{2}}\left(\bm{C}_{\gamma^{*}}(k)\right)
\end{equation}
Select $t_{2}$ as the prompted object $\bm{\tau}$, then $\mathcal{J}(\bm{\theta})$ is updated to:
\begin{equation}\label{eq:update_j}
\mathcal{J}(\bm{\theta})\!=\!f_{\hat{t}}(\bm{\theta})\!-\!f_{\bm{\tau}}(\bm{\theta})+\rho\left(1\!-\!f_{\bm{\tau}}(\bm{\theta}) / f_{\hat{t}}(\bm{\theta})\right)
\end{equation}
where $\rho$ is  {a} scale factor. This criterion  {reduces} the probability difference between the promoted class and the ground-truth class, increases the predicted probability of the promoted object, and speeds up the solution of attack parameters. The candidate population $\bm{C}(k)$ and the current population $\bm{X}(k)$ are judged by the above $\mathcal{J}(\bm{\theta})$, and the better individuals are selected to form the next generation $\bm{X}(k+1)$ (Step 14).
{For the impersonation attack, its solving objective is clear, that is, to raise the probability of the specified target identity. Therefore,}
the criterion is not adjusted (i.e. omit Step 8-13), and the solution is always along the direction of maximizing the probability of the target identity.

\subsection{The generation of adversarial stickers}\label{sec:transform}
After specifying the attack parameters, we deform the sticker accordingly to simulate the effect of the sticker on the face more realistically so that its shape fits the curvature of the face at the current position. We first use the 3DMM method \cite{blanz1999morphable} to generate a 3D model of a given 2D face image, and the 3D coordinates corresponding to the face position are obtained. Then we use the information of  {the} {X-Z} plane where the highest point $\left(x_{0}, y_{0}, z_{0}\right)$ of the pasting area is located to carry out  {the} bending transformation, and then use the information of  {the Y-Z} plane where $\left(x_{0}, y_{0}, z_{0}\right)$ is located to rotate the sticker in 3D space. The complete process of the shape transformation is shown in Figure \ref{fig:procedure}.

\begin{figure}[t]
\centering\includegraphics[width=0.42\textwidth]{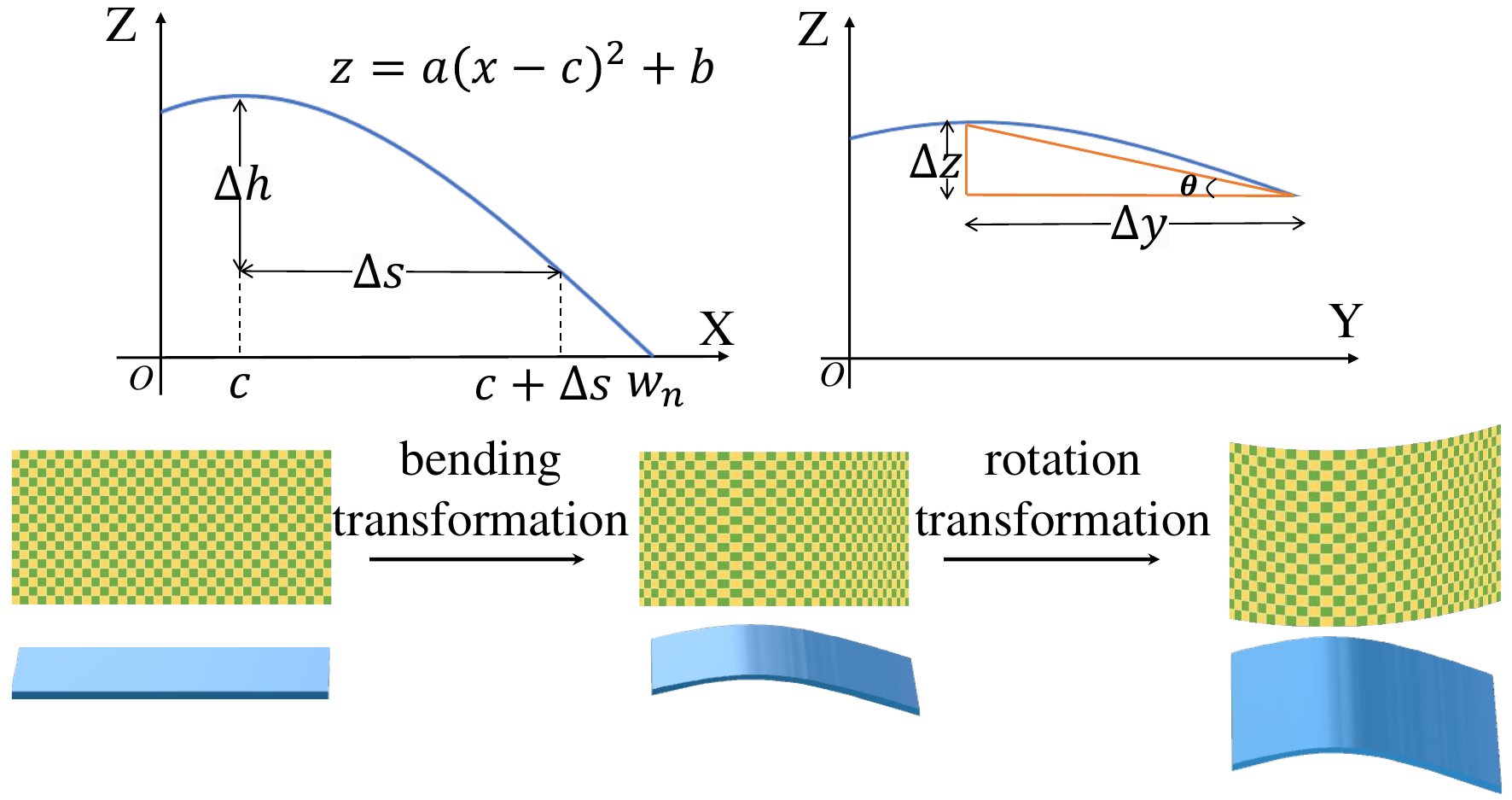}
\caption{The transformation process of the sticker pattern. The top row is the schematic diagram of variable annotation involved in  {the} deformation, the second row shows the change of sticker patterns on the 2D X-Y plane, and the third row is the stereogram of the corresponding sticker.}
\label{fig:transformation}
\end{figure}

For the bending transformation, the Y-axis coordinates of the sticker remain unchanged, we bend the sticker on the X-Z plane, and finally get the sticker $\bm{A}$ of size $h \times w_{n}$. The projection of points on the X-Z plane can be approximated as a parabola $z=a(x-c)^{2}+b$, where $c\!=\!x_{0}$, $a\!=\!-\Delta h /(\Delta s)^{2}$, $b\!=\!-a\left(w_{n}\!-\!c\right)^{2}$, $\Delta s$ is an arbitrary length and $\Delta h$ is the length on the Z-axis corresponding to $\Delta s$. The annotations of variables above and the visual effects are shown in Figure \ref{fig:transformation}. To ensure that the arc length of the bent sticker $\bm{A}$ is equal to the original width $w$, the width $w_{n}$ of $\bm{A}$ satisfies

\begin{equation}\label{eq:wn}
\int_{0}^{w_{n}} \sqrt{1+4 a^{2}(x-c)^{2}} d x=w
\end{equation}

Based on the parabola formula, we can get the matrix $M_{A} \in R^{ 3 \times \left(h * w_{n}\right)}$ containing the 3D coordinates of all pixels on  {the} sticker $\bm{A}$.

The bilinear interpolation and backward mapping are used to obtain the pixel value of each point in $\bm{A}$. Let $v_{p}(i, j)$ denote the pixel value at  {the} position $(i, j)$ in  {the} image $p$, then the pixel value on  {the} sticker $\bm{A}$ is calculated as follows

\begin{equation}\label{eq:va}
v_{A}(i, j)=g_{\bm{T}}\left(i, \int_{0}^{j} \sqrt{1+4 a^{2}(x-c)^{2}} d x\right)
\end{equation}where $i \in[0, h)$, $j \in[0, w_{n})$, $g_{p}(i, j)$ is used to calculate the pixel value corresponding to the position $(i,j)$ on the image $p$ after  {the} interpolation. Specifically,
\begin{equation}\label{eq:gp}
\begin{aligned}
g_{p}(u, v)=&\left.g_{p}(\lfloor u\rfloor+\Delta u, \lfloor v\rfloor+\Delta v\right) \\
=&(1-\Delta u)(1-\Delta v) \cdot v_{p}(\lfloor u\rfloor,\lfloor v\rfloor)+\\
& \Delta v(1-\Delta u) \cdot v_{p}(\lfloor u\rfloor,\lfloor v\rfloor+1)+\\
& \Delta u(1-\Delta v) \cdot v_{p}(\lfloor u\rfloor+1,\lfloor v\rfloor)+\\
&\left.\Delta u \Delta v \cdot v_{p}(\lfloor u\rfloor+1,\lfloor v\rfloor+1\right)
\end{aligned}
\end{equation}
where $\lfloor x\rfloor$ returns the largest integer less than or equal to $x$.

For  {the} rotation transformation, the information on the Y-Z plane reflects the rotation angle $\theta$ of the sticker, and $\theta$ of most of the sticker area is approximated to that of the entire sticker. $\Delta y$ denotes an arbitrary length and $\Delta z$ is the corresponding length on the Z-axis, then $\theta$ is calculated as follows:
\begin{equation}\label{eq:theta}
\theta=\operatorname{sign}\left(h-2 y_{0}\right) \cdot \arctan (\Delta z / \Delta y)
\end{equation}
where $\theta$, $\Delta y$, $\Delta z$, etc. are marked in Figure \ref{fig:transformation}. After the angle is calculated, the stickers after bending transformation are rotated in the 3-D space. Specifically, let $M_{B}$ denote the matrix formed by the coordinates of each point after rotation, then
\begin{equation}\label{eq:matrix}
M_{B}=\left[\begin{array}{ccc}
1 & 0 & 0 \\
0 & \cos \theta & -\sin \theta \\
0 & \sin \theta & \cos \theta
\end{array}\right] M_{A}
\end{equation}

According to the information of X and Y coordinates in $M_{B}$, the 2-D pattern of the sticker after  {the} complete deformation can be obtained by using the bilinear interpolation formulated in Eq. (\ref{eq:gp}).

\subsection{Implementation in the physical world}\label{sec:imp_phy}
Based on the above method, the attack parameters corresponding to the subjects' faces are solved in the digital environment. When conducting the physical attacks, we only need to paste the real stickers on the subjects' faces according to the calculated parameters. In this process, there are several points worth noting. (1) Our method does not involve the printing and making process, so there is no need to use  {the} NPS and TV losses with high calculation costs. (2) We do not use the EOT in the solving process to guarantee the performance under different physical conditions
but experiments in Section \ref{sec:physical} demonstrate that our method is robust under different physical conditions, such as changing face postures, which shows  {the} good adaptability of our method. (3) Even if there is a slight deviation between the calculated solution and the actual pasting position and angle, the follow-up experiments show that owing to the regional aggregation, it can still achieve good attacking results, verifying that the attack effectiveness caused by positions and rotation angles tends to keep consistent when the attacks are transferred to the physical environment.

\section{Extensions to other applications}
Besides  {the} face recognition, our method can be easily extended to other applications. In this section, we introduce two applications: image retrieval and traffic sign recognition.

\subsection{Image retrieval}
Image Retrieval (IR)  {system} returns a list of similar images sorted by  {the} similarity with the query image. Compared with adversarial attacks on  {the} face recognition task, which constructs adversarial  {patches} according to the returned labels and confidence scores, attacks for image retrieval are more challenging due to the similarity  {scores} between query  {images} and candidate images usually cannot be obtained in the black-box setting (such as Bing Image Search API, etc). To address this issue, we use the Relevance-based score proposed in \cite{li2021qair} as the similarity score, and then use the method in Section \ref{sec:method} to solve  {for} the available position and rotation angle of stickers.

Specifically, given a query image {$\bm{x}$} and a dataset $G$,  {the} image retrieval system $f$ will {return} top-$k$ images.
\begin{equation}
{RList_{k}(\bm{x},f)=\left\{\bm{x}_{1},...,\bm{x}_{i},...,\bm{x}_{k} | \bm{x}_{i}\in G \right\}}
\end{equation}
The returned images are sorted according to the similarity with  {the} query image $\bm{x}$, where $k$ is the number of images in the list. For most IR systems, because they usually return finite images, thus $k \ll |G|$.

To perform a successful attack, the objective function can be formulated as:
\begin{equation}\label{eq:ret_obj}
    RList_{k}(\bm{x},f) \cap RList_{k}(\bm{\tilde x} ,f)=\emptyset
\end{equation}
where $\bm{\tilde x}$ is the adversarial query image generated using the adversarial sticker described in the previous section.  

To solve the above equation, we use the relevance-based score \cite{li2021qair} to measure the attack effect:
\begin{equation}\label{eq:retrieval_loss}
    L(\bm{\tilde x},\bm{y})=\sum_{i=1}^{k}\omega_{i}\varphi_{i}
\end{equation}
where $\bm{\tilde x}=g(\bm{x} ; s, \bm{\theta})$ in Eq. (\ref{eq:untarget}), $\bm{y}=RList_{k}(\bm{x},f)$, and $\omega_{i}$ represents the ``prior sampling probability''\cite{li2021qair} and is given as:
\begin{equation}
    \omega_{i}=\frac{2^{r_{i}}-1}{\sum_{i=1}^{k}(2^{r_{i}}-1)}
\end{equation}
where $r_{i}=k-i$, $\varphi_{i}$ denotes ``attack failure probability" \cite{li2021qair} and is defined as follows:

\begin{equation}
    \varphi_{i}= \begin{cases}
    \omega_{i}, & \bm{x}_{i} \in RList_{k}(\bm{\tilde x},f). \\
    0, & \bm{x}_i \notin RList_{k}(\bm{\tilde x},f).
    \end{cases}
\end{equation}

\begin{figure}[h]
\centering\includegraphics[width=0.49\textwidth]{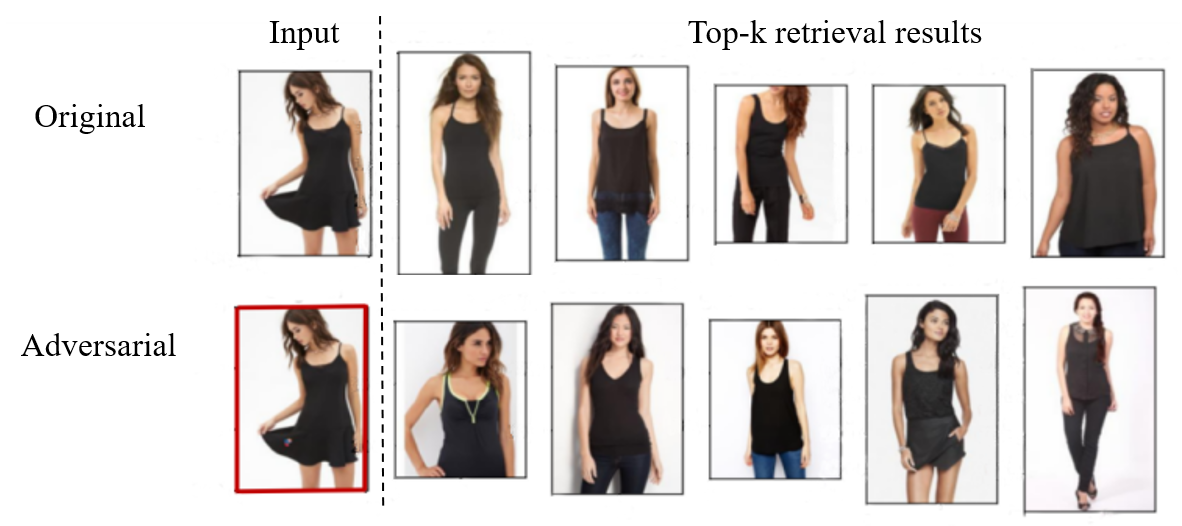}
\caption{{Failed attack results when $k$ is small (e.g. $k=5$). The} image in the red box is the adversarial example. We can find that the retrieved images are still relevant to adversarial query image after  {the} attack even {though} its original top-$k$ images are subverted.}
\label{fig:failure}
\end{figure}
{With the help of this defined relevance-based score, we can use $L(\bm{\tilde x},\bm{y})$ as $\mathcal{L}(\bm{\theta})$ in Algorithm \ref{alg:1}, making $L(\bm{\tilde x},\bm{y})$ as small as possible, to perform adversarial attacks.} If $L(\bm{\tilde x},\bm{y})=0$, Eq. (\ref{eq:ret_obj}) is achieved, and the attack is successful.
 However, we find it difficult to attack successfully when $k$ is small, and the attack effect is not obvious, which {can} be  {found} in Figure \ref{fig:failure}.

The main reason for the above phenomenon is {that the approach} we evaluate the attack effect can only confirm the empty set between the top-$k$ retrieved results before  and after attacks. As a large number of similar images appear in the dataset, we will still get  {the} secondary $k$ similar images  {to} the query image.  Therefore, the relevance-based score with a small $k$ is not sensitive enough to attacks, which could be not good for guiding to generate adversarial examples.
\begin{figure}[h]
\centering\includegraphics[width=0.49\textwidth]{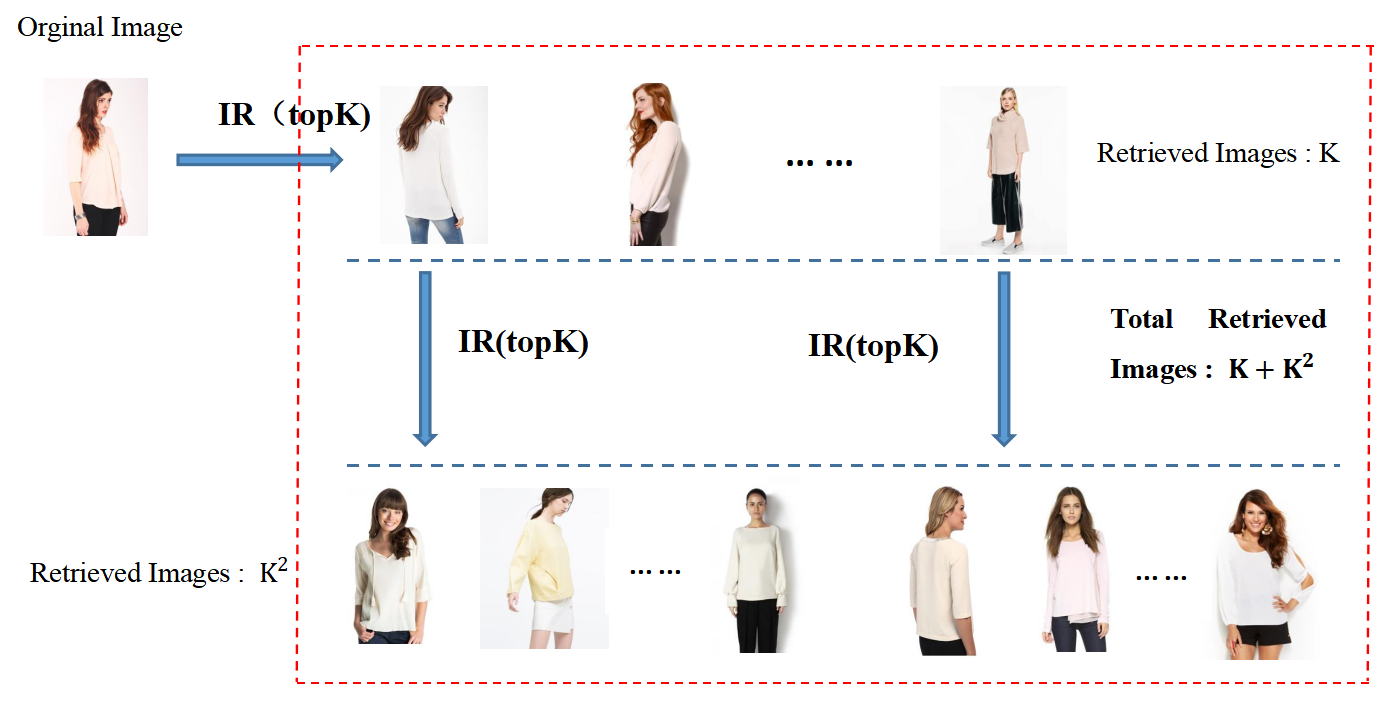}
\caption{Illustration for the proposed iterative query image retrieval.}
\label{fig:Iteration}
\end{figure}

Based on the above analysis, we decide to expand the retrieved images list of the query image before  {the} attack, so that our list contains enough images similar to the query image. Because $k$ is decided by the IR system, and cannot be changed by us. Instead, we propose an iterative query method to obtain more similar images. Specifically, we firstly get the top-$k$ retrieved images returned from  {the} IR system for the original query image, and then for each retrieved image, we search for its top-$k$ images again. Finally, all the above images ({at most} $k+k^2$ images) are regarded as candidate images.  The process is illustrated in Figure \ref{fig:Iteration}. In practice, we can firstly perform the attacks using the $k$ images output by  {the} IR system, if the attacking effect is satisfied, the attack task  is finished. If the attacking effect is not satisfied, we can expand the returned images using the iterative version, and judge the effect again.

\begin{table*}[t]
\caption{{The results of attacks on the face classification task.} We report the fooling rate (FR) and the number of queries (NQ) of the adversarial examples generated by different stickers on the LFW and CelebA datasets against FaceNet, SphereFace and CosFace. } 
\resizebox{\textwidth}{18.5mm}{
\begin{tabular}{c|c|cc|cc|cc|cc|cc|cc}
\hline
\multicolumn{2}{c|}{Datasets}               & \multicolumn{6}{c|}{LFW}                                                                      & \multicolumn{6}{c}{CelebA}                                                                  \\ \hline
\multicolumn{2}{c|}{\multirow{2}{*}{Model}} & \multicolumn{2}{c|}{FaceNet} & \multicolumn{2}{c|}{SphereFace} & \multicolumn{2}{c|}{CosFace} & \multicolumn{2}{c|}{FaceNet} & \multicolumn{2}{c|}{SphereFace} & \multicolumn{2}{c}{CosFace} \\ \cline{3-14} 
\multicolumn{2}{c|}{}  & {\small FR}              & {\small NQ}         & {\small FR}   & {\small NQ}     & {\small FR}   & {\small NQ}    & {\small FR}     & {\small NQ}   & {\small FR}    & {\small NQ}    & {\small FR}   & {\small NQ}   \\ \hline
\multirow{3}{*}{Dodging}  & sticker 1 & 63.22\%         & 489        & 42.74\%          & 691          & 54.28\%         & 527        & 73.51\%         & 518        & 57.18\%          & 596          & 69.47\%        & 530        \\
                                & sticker 2 & 76.26\%         & 478        & 64.08\%          & 629          & 69.82\%         & 484        & 81.78\%         & 483        & 72.93\%          & 576          & 79.26\%        & 487        \\
                                & sticker 3 & 73.64\%         & 442        & 44.50\%          & 604          & 66.59\%         & 455        & 80.33\%         & 511        & 59.92\%          & 548          & 72.80\%        & 496        \\ \hline
\multirow{3}{*}{Impersonation}  & sticker 1 & 51.11\%         & 636        & 30.70\%          & 718          & 48.06\%         & 563        & 48.18\%         & 610        & 37.32\%          & 644          & 42.90\%        & 653        \\
                                & sticker 2 & 50.00\%         & 715        & 31.00\%          & 870          & 45.93\%         & 658        & 48.96\%         & 652        & 41.67\%          & 747          & 47.73\%        & 637        \\
                                & sticker 3 & 46.28\%         & 691        & 29.50\%          & 716          & 45.54\%         & 662        & 47.84\%         & 625        & 39.18\%          & 700          & 45.83\%        & 638        \\ \hline
\end{tabular}}\label{tab:clf}
\end{table*}

\begin{figure*}[t]
\centering\includegraphics[width=0.95\textwidth]{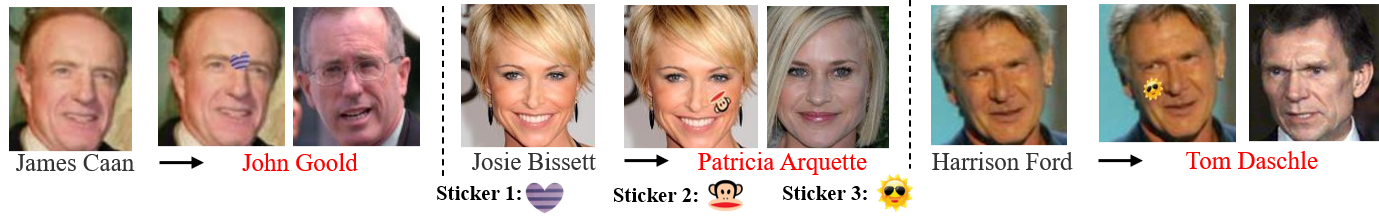}
\caption{Examples of attacks using different stickers. For each group, the three images correspond to the un-attacked original image, the image after attacks, and the image corresponding to the predicted wrong class after attacks. The black text denotes the predicted correct name and the red text denotes the predicted wrong name after attacks.}
\label{fig:adv_face1}
\end{figure*}

\subsection{Traffic sign recognition}
For the traffic sign recognition, attacks are extended to object detection models, which are used to detect and label multiple traffic signs within a scene \cite{song2018physical}. Instead of being limited to returning the class of  {a} single object in the image, the detector predicts both the locations (bounding boxes) and labels of multiple objects. Therefore, attacks in this task are more challenging. The attacks against object detectors can be divided into \textit{Disappearance Attacks} and \textit{Classification Attacks}. The goal of the former is to make the detector unable to detect the attacked sign (i.e. cannot return the bounding box corresponding to the object), while the latter focuses on modifying the label of the attacked detectable traffic sign.

Given an image with multiple traffic signs, assume  {the} object detector successfully detects and classifies  {the} traffic sign $\bm I$ and its bounding box is denoted as $O_{\textit{0}}$.  Our goal is to {paste} the adversarial sticker on this traffic sign to make the object detector predict wrong results. 
In each attack iteration,  because the image has multiple traffic {signs},  {the} object detector may output multiple bounding boxes $\bm{O}\!=\!\left\{O_{i} \mid i\!=\!1,\!\cdots,\! n\right\}$, we first need to align the target traffic sign $\bm I$. For that, we  find out the bounding box $O_{*}$ that has the largest IoU value \cite{yu2016unitbox} (i.e. the overlap area) with the reference box $O_{\textit{0}}$. This process can be expressed as:
\begin{equation}
O_{*}=\underset{i}{\arg \max } \ \operatorname{IoU}\left(O_{i}\right)=\underset{i}{\arg \max } \ \frac{\left|O_{0} \cap O_{i}\right|}{\left|O_{0} \cup O_{i}\right|}
\end{equation}
If IoU value \cite{yu2016unitbox} of $O_{*}$ is greater than the threshold $u$, $O_{*}$ is considered as the bounding box $O_{\bm I}$ corresponding to the traffic sign $\bm I$  in each attack iteration.

After aligning the bounding box in each attack iteration, we can obtain the score returned by  {the} object detector $f(\cdot)$ for the target traffic sign $\bm I$, that is:
\begin{equation}
f(\bm{x}, \bm{I}, t)=s\left(O_{\bm I}, t\right)
\end{equation}
where $s\left(O_{\bm I}, t\right)$ is the probability that the object corresponding to the bounding box $O_{\bm I}$ is predicted as the label $t$.

After obtaining the score of the attacked object, we can perform the classification attack according to the method described in Section \ref{sec:method} to modify the predicted label of $\bm I$. In the iterative process, if $\bm{O}$ is an empty set or no bounding box meets the threshold $u$, it is considered that the disappearance attack is realized.

\section{Experiments and Results}
\subsection{Experimental Settings}\label{sec:setting}
\noindent{\textbf{Target models}: We choose three representative face recognition models, CosFace \cite{wang2018cosface}, SphereFace \cite{liu2017sphereface}} and FaceNet \cite{schroff2015facenet}, as our target models.
{The open-source models\footnote{https://github.com/timesler/facenet-pytorch}\footnote{https://github.com/deepinsight/insightface/tree/master/recognition}\footnote{https://github.com/clcarwin/sphereface\_pytorch} are used to extract feature representations of faces.
For  {the} identification, we use the model to get the face embedding, and then take its nearest neighbor among all the identities in the dataset as the result of recognition.
For  {the} classification, we add the full connection layer after the above open-source model, and then fine-tune it on the corresponding datasets.
}
The dodging attack and the impersonation attack are conducted on all the above models.

\noindent{\textbf{Datasets}: We perform experiments on two public datasets: Labeled Faces in the Wild (LFW)\footnote{http://vis-www.cs.umass.edu/lfw/} and CelebFaces Attribute  (CelebA)\footnote{http://mmlab.ie.cuhk.edu.hk/projects/CelebA.html}. All 5749 identities of LFW and 8192 identities of CelebA are used to construct their own face databases. We select 1000 images randomly from each of the two datasets to carry out attacks. {Before attacks, we ensure that all the selected clean images can be correctly recognized by the model in both  {the} face classification and identification tasks.}}

\noindent{\textbf{Metrics}: Two metrics, Fooling Rate  {(FR)} and the Number of Queries  {(NQ)}, are used to evaluate the attack performance. The former refers to the percentage of all testing images that can be successfully attacked, while the latter refers to the number of model queries required for successful attacks. To study the effectiveness against face recognition modules, it is considered a successful attack if the face can successfully pass face detection and liveness detection but are identified as the wrong identity.}

\noindent{\textbf{Implementation}: We use {\small{\tt dlib}} library to extract 81 feature points of the face and fill the effective region to generate mask $M^{F}$. $d$ is equal to 2 in our case. $\theta_{1}$ refers to the index of the pasting position in the indexed set of valid points {\footnotesize$V\!:=\!\left\{(i, j) \mid M_{i j}^{F}\!=\!1\right\}$} and $\theta_{2}$ is the rotation angle. We set $r$ equal to 8. The default $l$ is equal to 1, and $l$ is increased if the corresponding point in the parameter space has already been accessed. 

 {In addition, we refer to the setting in \cite{one-pixel} and set $\alpha$ of Eq. (\ref{eq:cross}) to 0.5.
To ensure the balance between the two terms in Eq. (\ref{eq:update_j}), we set $\rho = 20$.
For the maximum number of iterations $T$, we set it to 30 based on the experience.
For the population size $P$, we set it to 80, 100, 120, and 140 respectively and perform some simple verifications, and find that when $P$ increases from 120 to 140, there is no obvious improvement in the success rate. Since the increase of $P$ will introduce a higher query cost, we set $P = 120$ in the subsequent experiments.
The threshold value $\delta$ is used to judge whether the probability difference reaches a small value, so we empirically set it to a small number (i.e. $\delta = 10$).}}

\subsection{Experimental Results}
\subsubsection{Performance comparisons in the digital world}
Firstly, we report the performance of our method on  {the} LFW and CelebA against FaceNet, SphereFace, and CosFace, respectively. We use three different stickers to conduct dodging and impersonation attacks under the face classification task and evaluate the fooling rate and the number of queries. The results are shown in Table \ref{tab:clf} and three groups of visual examples are given in Figure \ref{fig:adv_face1}. For the impersonation attack, we use the top-2 class for each face as the target class for simplicity.
\begin{table}[t]
\caption{{The distribution percentage of the three stickers in different positions when the attack is successful.}}
\centering  
\resizebox{0.48\textwidth}{7mm}{
\begin{tabular}{c|c|c|c|c|c}
\hline
         & Forehead & Between Eyebrows & Left Cheek & Right Cheek & Chin    \\ \hline
sticker1 & 17.39\%  & 28.50\%              & 19.42\%    & 20.56\%     & 14.13\% \\ \hline
sticker2 & 20.03\%  & 24.15\%              & 20.14\%    & 20.63\%     & 15.05\% \\ \hline
sticker3 & 20.11\%  & 27.36\%              & 20.49\%    & 21.23\%     & 10.81\% \\ \hline
\end{tabular}}\label{tab:st_post}
\end{table}

\begin{table}[t]
\caption{ Comparisons of the fooling rate and average time with two SOTA physical methods for face recognition systems in the black-box setting.}
\centering  
\resizebox{0.46\textwidth}{8mm}{
\begin{tabular}{c|c|c|c|c}
\hline
            & FaceNet           & SphereFace        & CosFace           & average time             \\ \hline
adv-hat     & 28.85\%           & 10.36\%           & 26.66\%           & 325.37s                 \\ \hline
adv-glasses & 21.21\%           & 10.63\%           & 9.83\%           & 536.25s                 \\ \hline
ours        & \textbf{76.26\%}  & \textbf{64.08\%}  & \textbf{69.82\%}  & \textbf{69.97s} \\ \hline
\end{tabular}}\label{tab:comparison}
\end{table}
From the above results, we can see: (1) The proposed Meaningful Adversarial Sticker method has shown good attack effectiveness in both dodging and impersonation attacks, achieving fooling rates of up to 81.78\% and 51.11\% respectively. (2) We can implement an attack at the magnitude of hundreds of queries, and,  {understandably, an} impersonation attack requires more queries than  {a} dodging attack, since the former requires perturbing the image to a specific class. (3) Under our attack, SphereFace shows strong robustness in both dodging and impersonation attacks, while FaceNet is relatively vulnerable.

{For different stickers, all of them can achieve successful attacks, but show different attack effects. Stickers with colorful patterns or some facial features (e.g. sticker 2 and sticker 3) show stronger attack effectiveness, especially in the case of  {a} dodging attack. We also make statistics on the paste positions of different stickers when successful attacks are achieved, as shown in Table \ref{tab:st_post}. It can be seen that when using the same sticker, the proportions of different positions do not show  {a} significant difference, and each position remains at about 20\%. But overall, the proportion is slightly higher between the eyes and slightly lower at the chin. Moreover, this phenomenon is consistent across different stickers.}

\subsubsection{Comparisons with SOTA methods}
\begin{figure}[t]
\centering\includegraphics[width=0.4\textwidth]{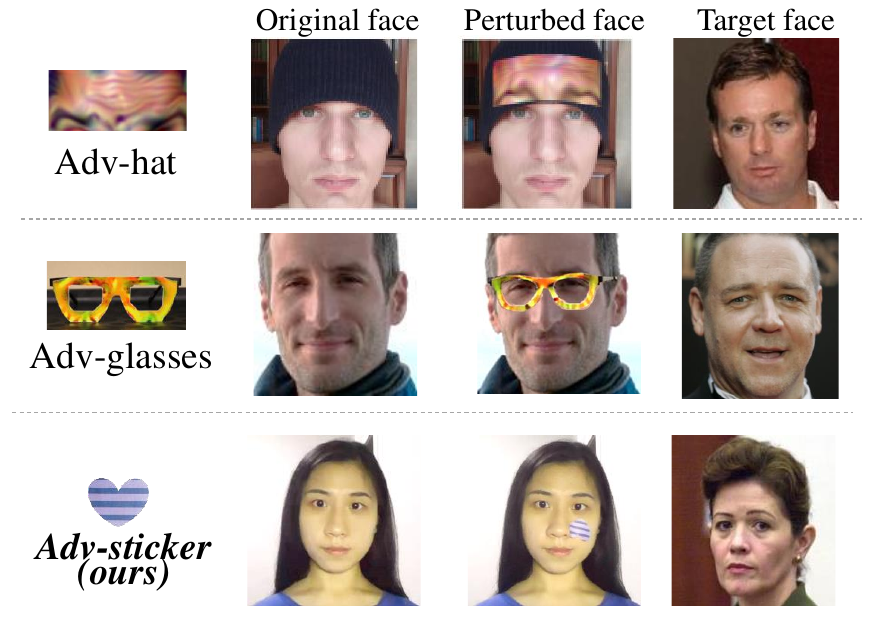}
\caption{{Comparisons of our meaningful Adv-sticker with other attack methods (Adv-hat \cite{komkov2019advhat}, Adv-glasses \cite{sharif2016accessorize,sharif2019general}) for FR systems. Our approach uses real stickers without relying on the generated perturbation patterns and printed accessories.}}
\label{fig:figure1}
\end{figure}

Comparisons between our method and other physically realizable attacks for  {the} face recognition on the same face images in the LFW are shown in Table \ref{tab:comparison}. Since there are no existing physical attacks on  {the} face recognition in the black-box setting, we can only use the adversarial examples generated in  {the} white-box setting to carry out transfer-based black-box attacks when calculating the performance of the previous methods. Although score-based black-box attacks can also be carried out through  {the} gradient estimation, it is not realistic because the estimation of large areas of pixel gradients requires a large number of model queries.
Here we choose adv-hat \cite{komkov2019advhat} and adv-glasses \cite{sharif2016accessorize} which have great performance on the white-box setting. 
For our method, we use the results of dodging  {attacks} by pasting sticker 2 to compare with other methods. The results in Table \ref{tab:comparison} show that our method can achieve better attack effectiveness in a shorter time when attacking different networks. It outperforms adv-hat with at most more than 83\% improvement and adv-glasses with at most 85\% improvement, while the average time to attack each image is reduced by 78\% and 86\% respectively.
The qualitative {comparisons} between our method and adv-hat \cite{komkov2019advhat} and adv-glasses \cite{sharif2016accessorize}  are given in Figure \ref{fig:figure1}.

\begin{figure}[t]
\centering\includegraphics[width=0.48\textwidth]{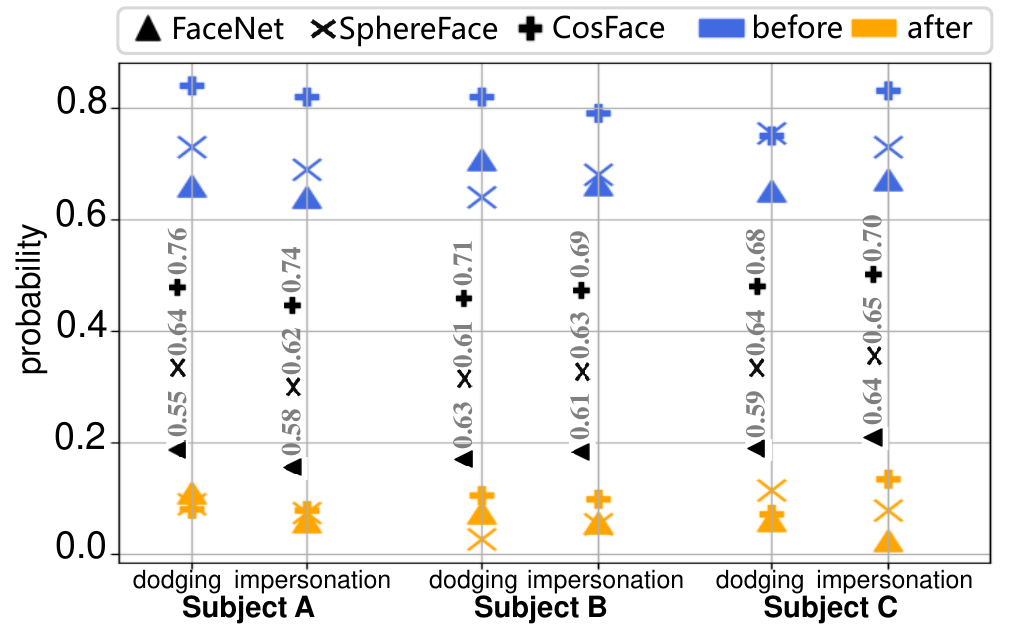}
\caption{The predicted probabilities of ground-truth labels before and after attacks in  {the} physical environment. The numbers next to the vertical line represent the difference in  {the} probability on each model.}
\label{fig:deltap}
\end{figure}

\begin{table}[t]
\caption{{The fooling rate (FR) and the number of queries (NQ) of  {the} ablation study. Results are obtained on  {the} LFW dataset to carry out dodging attacks using sticker 2.}}
\centering  
\resizebox{0.48\textwidth}{14mm}{
\begin{tabular}{c|p{0.9cm}<{\centering}p{0.7cm}<{\centering}|p{0.9cm}<{\centering}p{0.7cm}<{\centering}|p{0.9cm}<{\centering}p{0.6cm}<{\centering}}
\hline
\multirow{2}{*}{} & \multicolumn{2}{c|}{FaceNet} & \multicolumn{2}{c|}{SphereFace} & \multicolumn{2}{c}{CosFace} \\ \cline{2-7} 
                  & {\small FR}             & {\small NQ}          & {\small FR}                & {\small NQ}           & {\small FR}               & {\small NQ}        \\ \hline
DE                & 29.03\%        & 1107        & 21.88\%          & 1262         & 31.83\%         & 768        \\ \hline
adaptive-DE       & 41.96\%        & 764         & 34.38\%          & 862          & 38.12\%         & 564        \\ \hline
region-DE         & 54.81\%        & 871         & 41.67\%          & 970          & 55.59\%         & 519        \\ \hline
{RHDE {\small(ours)}}  & 76.26\%        & 478        & 64.08\%         & 629          & 69.82\%         & 484        \\ \hline \hline
upper bound       & 79.52\%        & 11640       & 67.31\%          & 13128        & 73.89\%         & 12804        \\ \hline
\end{tabular}}\label{tab:ablation}
\end{table}

\subsubsection{Ablation study}
In this section, we first demonstrate the effectiveness of each component in the proposed method, and report the performance when each component of our RHDE algorithm is added separately.
We conduct experiments on  {the} LFW dataset to
carry out dodging attacks using sticker 2.
In all experiments, the population size $P$ and iteration number $T$ are consistent. Starting from the traditional differential evolution algorithm (DE), we add  {the} adaptive adjustment strategy (adaptive-DE) and  {the} region-based offspring generation strategy (region-DE), respectively, The comparison results are shown in Table \ref{tab:ablation}.

\begin{table}[t]
\caption{The percentage of video frames successfully attacked when different subjects continuously change their face postures in the physical environment.}
\centering  
\begin{threeparttable}
\resizebox{0.46\textwidth}{56.0mm}{
\begin{tabular}{c|c|c|c|c}
\hline
model                       & subject & with-def             & no-def    & difference \\ \hline
\multirow{9}{*}{FaceNet}    & A       & \textbf{98.46\%}     & 59.78\%   & 38.68\%                \\ \cline{2-5} 
                            & B       & \textbf{98.30\%}     & 48.74\%   & 49.56\%               \\ \cline{2-5} 
                            & C       & \textbf{92.94\%}     & 48.39\%   & 44.55\%               \\ \cline{2-5} 
                            & D       & \textbf{97.50\%}     & 53.87\%   & 43.63\%               \\ \cline{2-5} 
                            & E       & \textbf{98.16\%}     & 58.96\%   & 39.20\%               \\ \cline{2-5} 
                            & F       & \textbf{95.36\%}     & 48.25\%   & 47.11\%               \\ \cline{2-5} 
                            & G       & \textbf{97.87\%}     & 51.16\%   & 46.71\%               \\ \cline{2-5} 
                            & H       & \textbf{98.24\%}     & 57.85\%   & 40.39\%               \\ \cline{2-5} 
                            & I       & \textbf{96.97\%}     & 55.73\%   & 41.24\%                \\ \hline
\multirow{9}{*}{SphereFace} & A       & \textbf{91.30\%}     & 55.17\%   & 36.13\%                \\ \cline{2-5} 
                            & B       & \textbf{83.45\%}     & 33.33\%   & 50.12\%               \\ \cline{2-5} 
                            & C       & \textbf{85.92\%}     & 40.76\%   & 45.16\%               \\ \cline{2-5} 
                            & D       & \textbf{83.87\%}     & 40.88\%   & 42.99\%               \\ \cline{2-5} 
                            & E       & \textbf{84.92\%}     & 41.24\%   & 43.68\%               \\ \cline{2-5} 
                            & F       & \textbf{89.76\%}     & 48.60\%   & 41.16\%               \\ \cline{2-5} 
                            & G       & \textbf{90.45\%}     & 49.71\%   & 40.74\%               \\ \cline{2-5} 
                            & H       & \textbf{86.90\%}     & 44.54\%   & 42.36\%               \\ \cline{2-5} 
                            & I       & \textbf{83.48\%}     & 42.93\%   & 40.55\%                \\ \hline
\multirow{9}{*}{CosFace}    & A       & \textbf{85.37\%}     & 48.09\%   & 37.28\%                \\ \cline{2-5} 
                            & B       & \textbf{86.96\%}     & 46.67\%   & 40.29\%               \\ \cline{2-5} 
                            & C       & \textbf{82.61\%}     & 45.45\%   & 37.16\%               \\ \cline{2-5} 
                            & D       & \textbf{84.39\%}     & 44.96\%   & 39.43\%               \\ \cline{2-5} 
                            & E       & \textbf{86.06\%}     & 47.61\%   & 38.45\%               \\ \cline{2-5} 
                            & F       & \textbf{83.42\%}     & 41.79\%   & 41.63\%               \\ \cline{2-5} 
                            & G       & \textbf{84.88\%}     & 45.82\%   & 39.06\%               \\ \cline{2-5} 
                            & H       & \textbf{82.29\%}     & 43.60\%   & 38.69\%               \\ \cline{2-5} 
                            & I       & \textbf{85.91\%}     & 48.95\%   & 36.96\%                \\ \hline
\end{tabular}}\label{tab:frame}
\begin{tablenotes}
\item\footnotesize\emph{with-def}: using 3d deformation in parameters' solving process. \\\emph{no-def}: pasting the sticker directly without considering deformation. \\\emph{difference}: the difference between with-def and no-def.
\end{tablenotes}
\end{threeparttable}
\end{table}

We can see that under the same maximum number of iterations, the  {fooling} rates of directly using DE are very low and relatively more queries are required. When the adaptive adjustment strategy and the offspring's generation strategy are added respectively, the  {fooling} rates of both are improved. When the two strategies are used together, the  {fooling} rates are greatly improved, and  {the} queries required are significantly reduced.

\begin{figure*}[t]
\centering\includegraphics[width=0.98\textwidth]{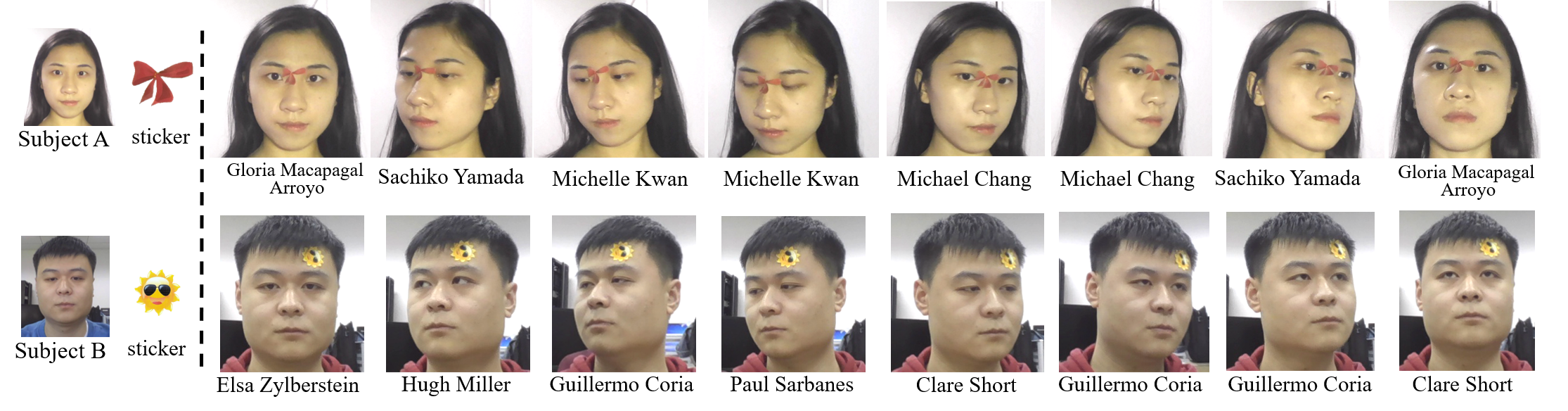}
\caption{ Examples showing the attack effectiveness at different face postures in the physical environment (un-targeted attacks). The black text on the right side denotes the predicted wrong name after attacks.}
\label{fig:frame}
\end{figure*}
{We also conduct experiments on the upper bound of the achievable  {fooling} rate.
We take 30 different angles of stickers at an interval of 12 degrees, consider every valid paste position of the face, and traverse all possible solutions in the search space. When a successful attack is implemented, the traversal is stopped. The corresponding results of dodging attacks using sticker 2 on  {the} LFW dataset are shown in the ``upper bound'' of Table \ref{tab:ablation}. It can be seen that the upper bound of the face recognition on these three models is only 3.26\%, 3.23\%, and 4.07\% higher than our method, but NQ is 11162, 12499, and 12320 more than our method. This proves that our method can efficiently find the successful solution.}

\subsubsection{Attacks in the physical world}\label{sec:physical}
In this section, we report the performance of our meaningful adversarial stickers in the physical environment.
Figure \ref{fig:deltap} presents the predicted probabilities of three subjects corresponding to the ground-truth identity before and after attacks in the physical environment. The results show that the probabilities in different models are significantly reduced, and the maximum reduction in FaceNet, SphereFace, and CosFace after attacks are 0.64, 0.65, and 0.76, respectively. This proves that the generated attack parameters in the digital environment can still maintain  {a} good attack performance when applied to the physical world.

We also report the results of  {fooling} rates in complex physical conditions. We use the parameters calculated in the digital world, change face postures (counterclockwise rotation of the head) in the physical world, and count the percentage of successful attacks in consecutive frames. To prove the necessity of  {the} 3D deformation (described in Section \ref{sec:transform}), we also calculate the relevant physical results of the parameters generated when 3D deformation is not considered. Table \ref{tab:frame} shows the above results for some subjects and Figure \ref{fig:frame} shows the visual examples of our method at different face postures.

It demonstrates that our method still has {a} good attacking performance when changing face postures in the physical world. If the curvature of the human cheek is ignored and  {the} 3D deformation is not considered, the corresponding performance in the physical environment will be greatly weakened. Importantly, it can maintain such good attack effectiveness in the physical world without considering complex physical conditions when solving parameters.

\begin{table}[t]
\caption{ {The fooling rate (FR) and the number of queries (NQ) after adversarial training, and the changes compared to the undefended situations (shown in brackets). Here we use sticker 2 to conduct dodging attacks on LFW.}}
\centering  
\resizebox{0.5\textwidth}{10mm}{
\begin{tabular}{p{0.83cm}<{\centering}|p{0.32cm}<{\centering}|p{0.73cm}<{\centering}p{1.3cm}<{\raggedright}|p{0.73cm}<{\centering}p{1.3cm}<{\raggedright}|p{0.73cm}<{\centering}p{1.3cm}<{\raggedright}}
\hline
training                       &    & \multicolumn{2}{c|}{FaceNet} & \multicolumn{2}{c|}{SphereFace} & \multicolumn{2}{c}{CosFace} \\ \hline
\multirow{2}{*}{\small sticker1}    & {\small FR} & 74.18\%  & ($\downarrow$2.08\%)   & 61.02\%    & ($\downarrow$3.06\%)     & 67.46\%  & ($\downarrow$2.36\%)      \\ \cline{2-8} 
                        & {\small NQ} & 485      & ($\uparrow$ 7)        & 646        & ($\uparrow$ 17)           & 498         & ($\uparrow$ 14)     \\ \hline
\multirow{2}{*}{\small sticker2} & {\small FR} & 60.35\%  & ($\downarrow$15.91\%)   & 47.29\%    & ($\downarrow$16.79\%)     & 51.09\%      & ($\downarrow$18.73\%)       \\ \cline{2-8} 
                        & {\small NQ} & 532     & ($\uparrow$ 54)        & 675       & ($\uparrow$ 46)         & 544      & ($\uparrow$ 60)            \\ \hline
\end{tabular}
}\label{tab:defense}
\end{table}

\subsubsection{Robustness of meaningful adversarial stickers}

We also test the robustness of our method in response to defense measures. We here choose adversarial training \cite{madry2018towards} as the defense method and conduct experiments on the LFW dataset.
 {Specifically, we train two kinds of robust models using the generated adversarial examples with the sticker 1 and sticker 2 in Figure \ref{fig:adv_face1}, respectively. And then the adversarial sticker attacks are conducted against these robust models using the sticker 2}. Table \ref{tab:defense} lists the fooling rate and the number of queries of dodging attacks after the defense with sticker 2, as well as the changes compared to the results without defense.
 {Generally speaking, the fooling rate will decrease and the query number will increase because of the improved robustness after the adversarial training. It can be seen that if we use the same sticker for the models' training and subsequent attacks, adversarial training does cause a drop in the fooling rate (i.e. the last two lines of Table \ref{tab:defense}). However, if the stickers used for training (sticker 1) and subsequent attacks (sticker 2) are different, the variation range of fooling rates and queries is relatively small, with the maximum variation range of 3.06\% and 17 respectively. This shows that only a sticker is ``seen" in the training set, adversarial training can resist the attacks caused by this sticker. However, in practice, the sticker types are so wide that defenders often have no way of knowing which sticker patterns are used by attackers. Therefore, it is more common for adversarial training and attacks to use different stickers. So in real applications, our attack method has a good robustness against the adversarial training.}

\begin{figure}[t]
\centering\includegraphics[width=0.4\textwidth]{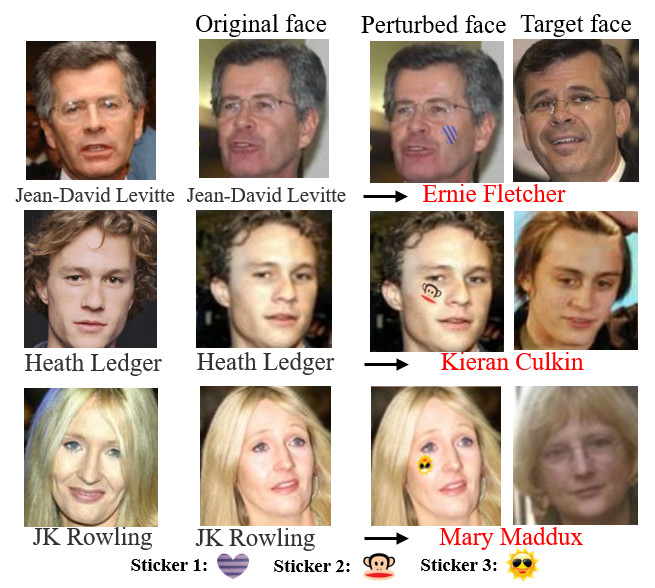}
\caption{Examples using different stickers under  {the} face identification task in the digital world. The black text denotes the predicted correct name and the red text denotes the predicted wrong name after attacks.}
\label{fig:adv_face}
\end{figure}

\begin{table*}[t]
\caption{{The results of attacks on the face identification task. We report the fooling rate (FR) and the number of queries (NQ) of the adversarial examples generated by different stickers on the LFW and CelebA datasets against FaceNet and the commercial API service (cml. API).}}
\centering  
\begin{tabular}{p{0.9cm}<{\centering}|p{0.77cm}<{\centering}p{0.16cm}<{\centering}p{0.77cm}<{\centering}p{0.16cm}<{\centering}p{0.77cm}<{\centering}p{0.16cm}<{\centering}p{0.77cm}<{\centering}p{0.35cm}<{\centering}|p{0.77cm}<{\centering}p{0.16cm}<{\centering}p{0.77cm}<{\centering}p{0.16cm}<{\centering}p{0.77cm}<{\centering}p{0.16cm}<{\centering}p{0.77cm}<{\centering}p{0.16cm}<{\centering}}
\hline
\multirow{4}{*}{} & \multicolumn{8}{c|}{Dodging}                                                                                                                 & \multicolumn{8}{c}{Impersonation}                                                                                                           \\ \cline{2-17} 
                  & \multicolumn{4}{c|}{LFW}                                                & \multicolumn{4}{c|}{CelebA}                                        & \multicolumn{4}{c|}{LFW}                                                & \multicolumn{4}{c}{CelebA}                                        \\ \cline{2-17} 
                  & \multicolumn{2}{c|}{FaceNet}       & \multicolumn{2}{c|}{cml. API}      & \multicolumn{2}{c|}{FaceNet}       & \multicolumn{2}{c|}{cml. API} & \multicolumn{2}{c|}{FaceNet}       & \multicolumn{2}{c|}{cml. API}      & \multicolumn{2}{c|}{FaceNet}       & \multicolumn{2}{c}{cml. API} \\ \cline{2-17} 
                  & FR      & \multicolumn{1}{c|}{NQ}  & FR      & \multicolumn{1}{c|}{NQ}  & FR      & \multicolumn{1}{c|}{NQ}  & FR              & NQ          & FR      & \multicolumn{1}{c|}{NQ}  & FR      & \multicolumn{1}{c|}{NQ}  & FR      & \multicolumn{1}{c|}{NQ}  & FR              & NQ         \\ \hline
sticker1          & 43.11\% & \multicolumn{1}{c|}{504} & 39.05\% & \multicolumn{1}{c|}{524} & 66.44\% & \multicolumn{1}{c|}{574} & 43.95\%         & 513         & 41.63\% & \multicolumn{1}{c|}{586} & 34.65\% & \multicolumn{1}{c|}{594} & 58.82\% & \multicolumn{1}{c|}{552} & 37.96\%         & 581        \\
sticker2          & 43.68\% & \multicolumn{1}{c|}{484} & 42.81\% & \multicolumn{1}{c|}{492} & 75.41\% & \multicolumn{1}{c|}{472} & 46.17\%         & 489         & 42.11\% & \multicolumn{1}{c|}{596} & 35.24\% & \multicolumn{1}{c|}{575} & 55.87\% & \multicolumn{1}{c|}{511} & 40.25\%         & 572        \\
sticker3          & 48.82\% & \multicolumn{1}{c|}{422} & 40.76\% & \multicolumn{1}{c|}{485} & 80.03\% & \multicolumn{1}{c|}{439} & 45.50\%         & 467         & 46.85\% & \multicolumn{1}{c|}{587} & 35.93\% & \multicolumn{1}{c|}{583} & 55.15\% & \multicolumn{1}{c|}{563} & 39.51\%         & 590        \\ \hline
\end{tabular}\label{tab:idt}
\end{table*}

\subsection{Results of the face identification task}
Besides face classification models mentioned above, there is another face recognition task called face identification models widely used in our life. Different from  {the} face classification, where the model outputs the identity label and probability with a  classifier based on the extracted features,   {the} face identification uses the face feature representation obtained by the models (CosFace, SphereFace, and FaceNet, etc) to calculate  {the} cosine similarity with all face images in the database, and takes the class with the highest similarity as the predicted identity, which belongs to the category of  {the} metric learning.

\begin{figure}[t]
\centering\includegraphics[width=0.49\textwidth]{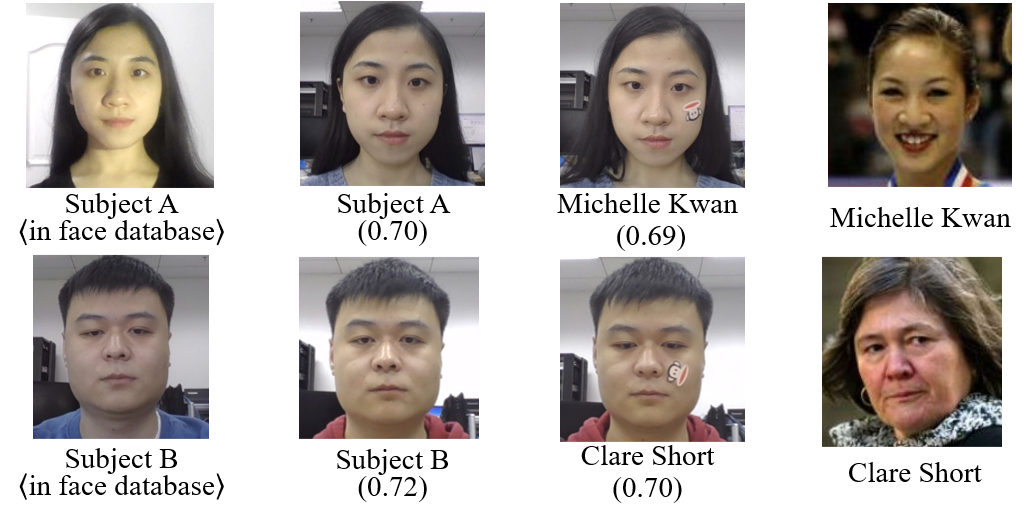}
\caption{Examples of attacks using different stickers under the face identification task in the physical world. The black text at the bottom of the image denotes the identified  {person's} name and the corresponding cosine similarity (shown in brackets).}
\label{fig:phy_verify}
\end{figure}

\begin{figure}[t]
\centering\includegraphics[width=0.49\textwidth]{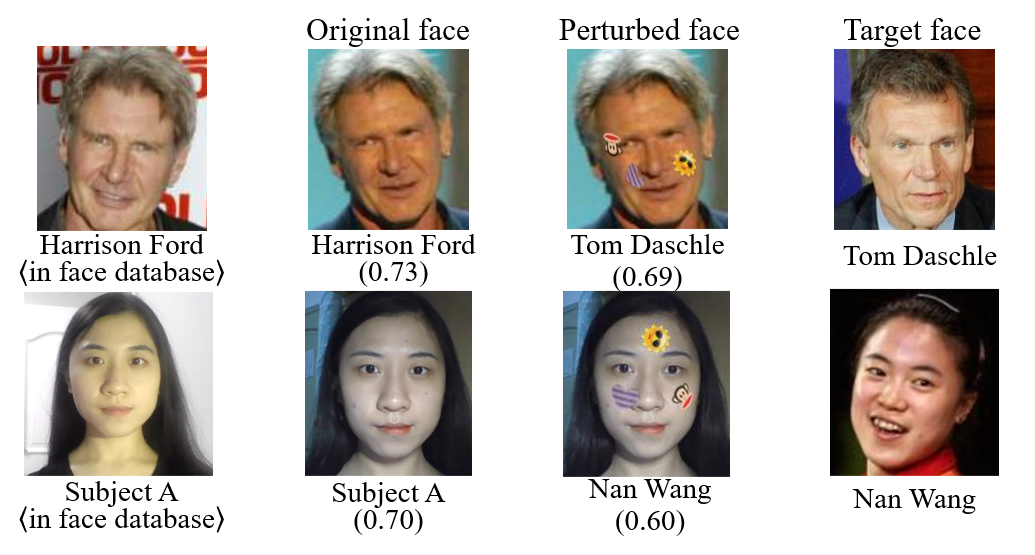}
\caption{Examples using multiple stickers under the face identification task. The black text at the bottom of the image denotes the identified  {person's} name and the corresponding cosine similarity (shown in brackets).}
\label{fig:multiple6}
\end{figure}

In this subsection, we show the corresponding results about  {the} face identification task.
{In addition to the open-source model FaceNet, we also conduct experiments on the commercial face recognition API service\footnote{https://intl.cloud.tencent.com/product/facerecognition} to verify the effectiveness of our method in the practical scenario.}
The results of using different stickers to attack faces in the LFW and CelebA datasets are shown in Table \ref{tab:idt} and Figure \ref{fig:adv_face} presents some examples in the digital world. It can be seen that in the face identification task, our attack method can still maintain  {the} good attack performance with  {a} natural appearance in both dodging attack and impersonation attacks, achieving the fooling rate of at most 80.03\% and 58.82\%, respectively.
{It can also reach 46.17\% and 40.25\% on the commercial API. Although the performance is slightly lower than that of the open-source model, it is still acceptable because the commercial system includes some defense mechanisms such as the image compression.}
Besides, in most cases, face identification based on  {the} metric learning shows better robustness than face classification based on  {the} traditional model classification. This phenomenon is worthy of our further exploration. Figure \ref{fig:phy_verify} gives some examples of physical attacks under  {the} face identification task. 

We also show the results of using multiple stickers to attack the face identification task. Figure \ref{fig:multiple6} shows some visual examples. The results show that when using multiple stickers to attack (the stickers do not cover the key features of the face and do not overlap each other), it can still achieve a good attack performance.

\begin{table}[t]
\caption{ The attack results for image retrieval using adversarial stickers.}
\centering  
\begin{tabular}{c|c|c|c|c|c||c}
\hline
 Recall@K      & 1            & 10         & 20          & 30          &40          & NQ              \\ \hline
Original            & 0.642    & 0.868	& 0.910	& 0.926	& 0.938	& 0	             \\ \hline
Adversarial     &0.068	&0.214	&0.296	&0.346     &0.360	&343	              \\ \hline
\end{tabular}\label{tab:Performance}
\end{table}


\begin{figure}[t]
\centering\includegraphics[width=0.49\textwidth]{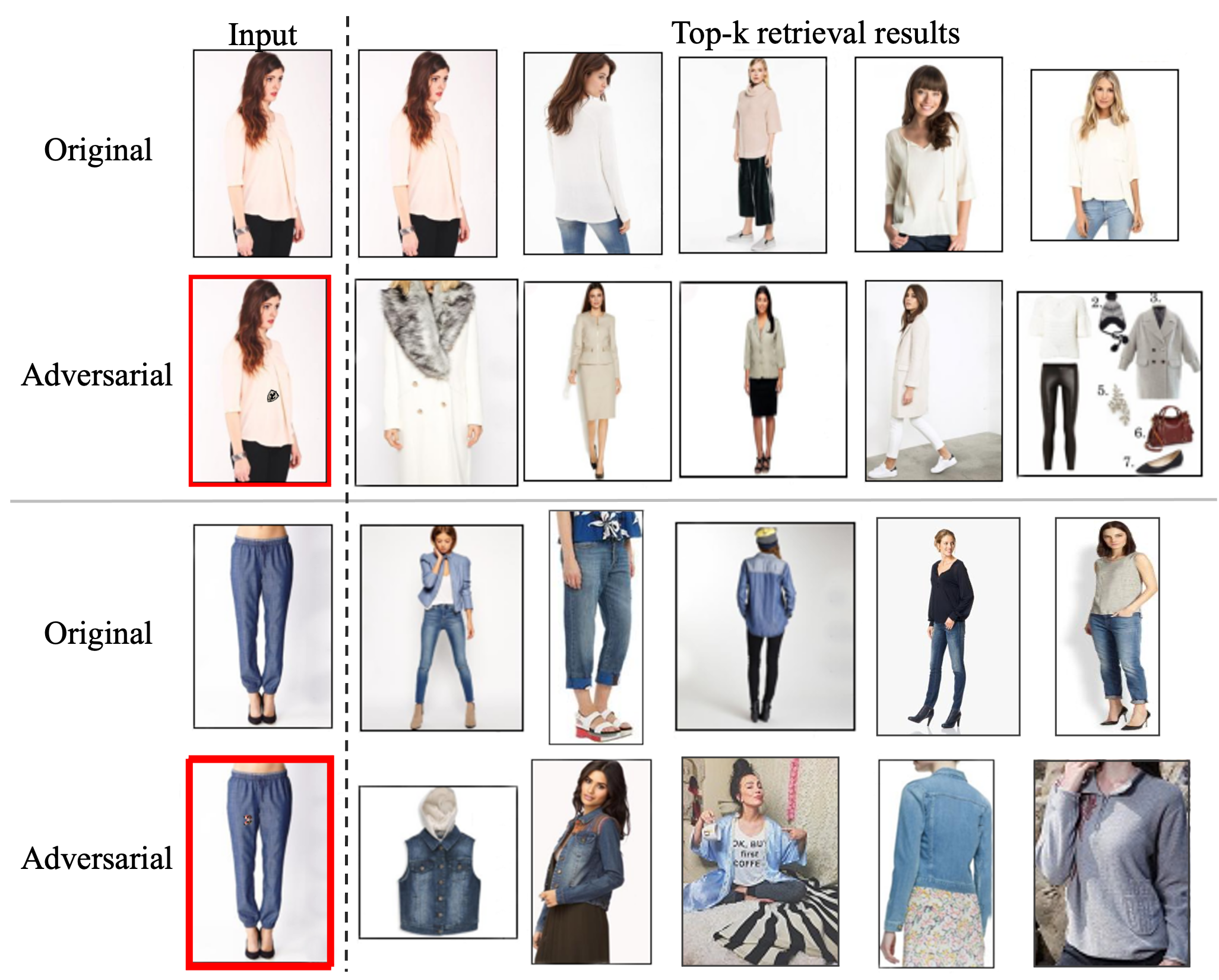}
\caption{{ {Two groups of attack results for the image retrieval system. The image in the red box is the adversarial example. We can find that the returned clothes after the attack are not related to the original results.}}}
\label{fig:example}
\end{figure}

\section{Extensions to other applications}

In this section, we introduce the corresponding experimental results versus  {the} traffic sign recognition task and image retrieval task. 

\subsection{Image retrieval}

To attack  {the} image retrieval system, we choose the In-shop clothes retrieval task \cite{liuLQWTcvpr16DeepFashion}, which is to determine if two images taken in  {the} shop belong to the same clothing item. The used dataset  \cite{liuLQWTcvpr16DeepFashion} contains 54, 642 images of 11, 735 clothing items. Top-$k$ retrieval accuracy (recall@K) is adopted to measure the performance of fashion retrieval.
We use the deep-fashion-retrieval\footnote{https://github.com/ihciah/deep-fashion-retrieval} as the target image retrieval system to attack.

The quantitative attack performance is given in Table \ref{tab:Performance}. From the table, we can see that for the same image retrieval system,  {the} adversarial query image obtains lower Recall@K than  {the} original query image. The gap is nearly 0.6.  The number of query times is 343. This table shows that our method can use a few query times to achieve a big performance drop. 

\begin{table}[t]
\caption{ The quantitative attack performance for traffic sign recognition using  {an} adversarial sticker.}
\centering  
\begin{tabular}{c|c|c}
\hline
           & Original  & Adversarial                     \\ \hline
 mAP  & 0.82 	   & 0.23		                \\ \hline
 FR     & 0            &83.5\%                             \\ \hline
 NQ    & 0            &435                                     \\ \hline
\end{tabular}\label{tab:Performance34}
\end{table}

Two attacking examples are shown in Figure \ref{fig:example}. From the figure, we can see that the image system can successfully search for the similar clothes with the original query cloth (the first row and the third row), but when the sticker is pasted on the query cloth, the returned clothes by  {the} image retrieval system are  {weakly correlated with the given input} (the second row and fourth row).  {For example, after putting the sticker on the jeans, the retrieval system returns some jackets' images}. The pasted sticker is small compared with the cloth, but it will affect the results output by the image retrieval system, which shows the weak robustness of  {the} image retrieval system against adversarial attacks.

\begin{figure}[t]
\centering\includegraphics[width=0.49\textwidth]{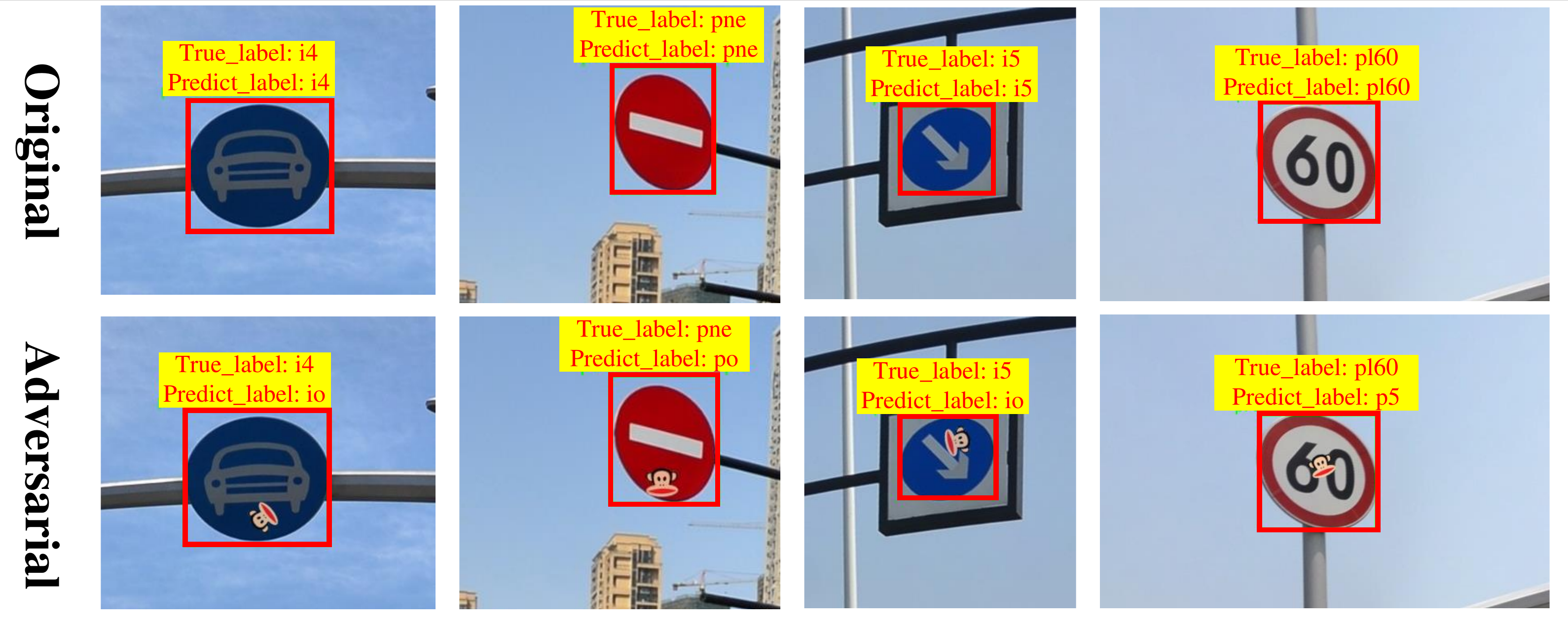}
\caption{Classification Attack for traffic sign recognition in the digital world.  {The top row denotes the results on the original images, and the bottom row denotes the results on the adversarial images}.}
\label{fig:multiple1}
\end{figure}
\begin{figure}[t]
\centering\includegraphics[width=0.488\textwidth]{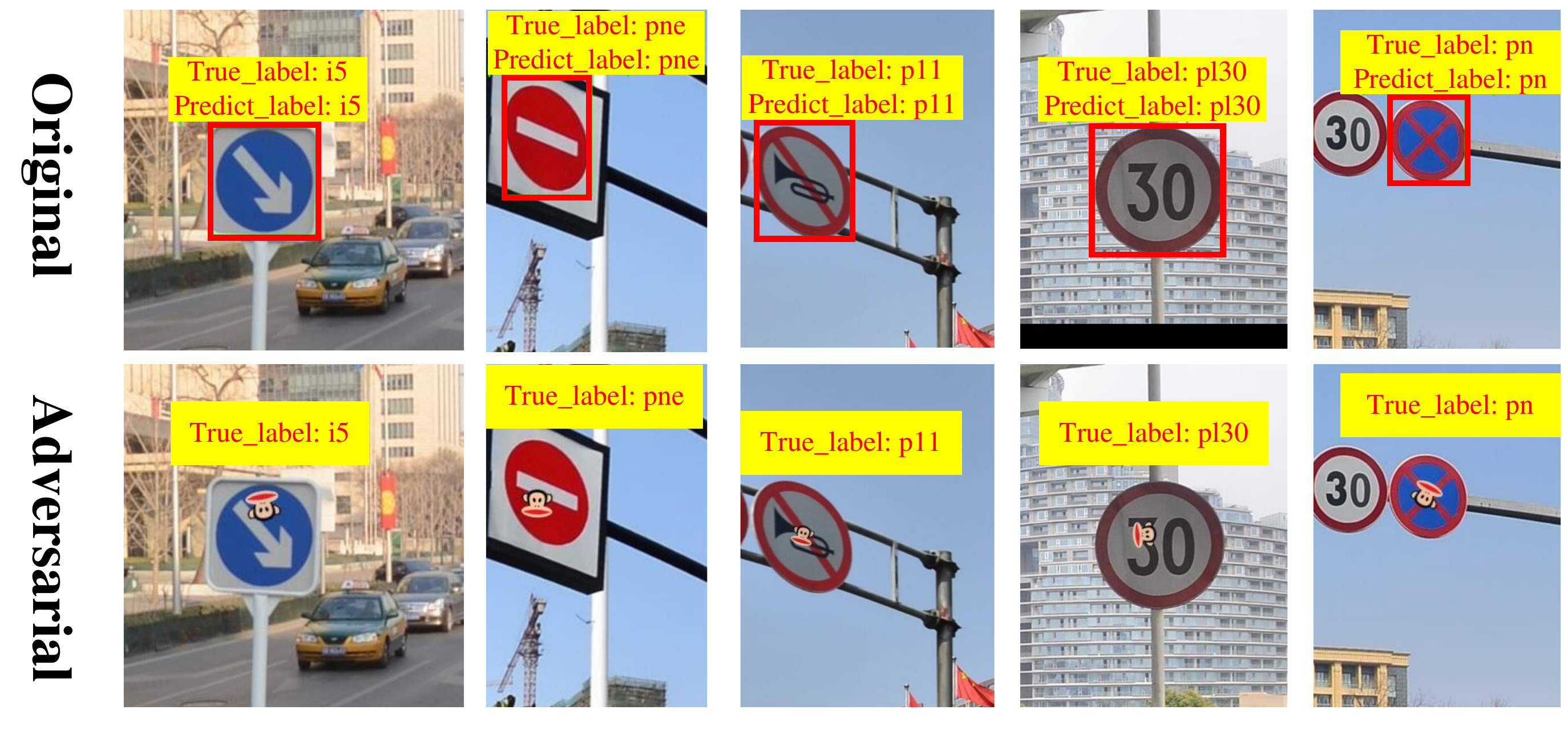}
\caption{Disappearance attack for traffic sign recognition in the digital world.  {The top row denotes the results on the original images, and the bottom row denotes the results on the adversarial images}.}
\label{fig:multiple2}
\end{figure}

\begin{figure*}[t]
\centering\includegraphics[width=0.95\textwidth]{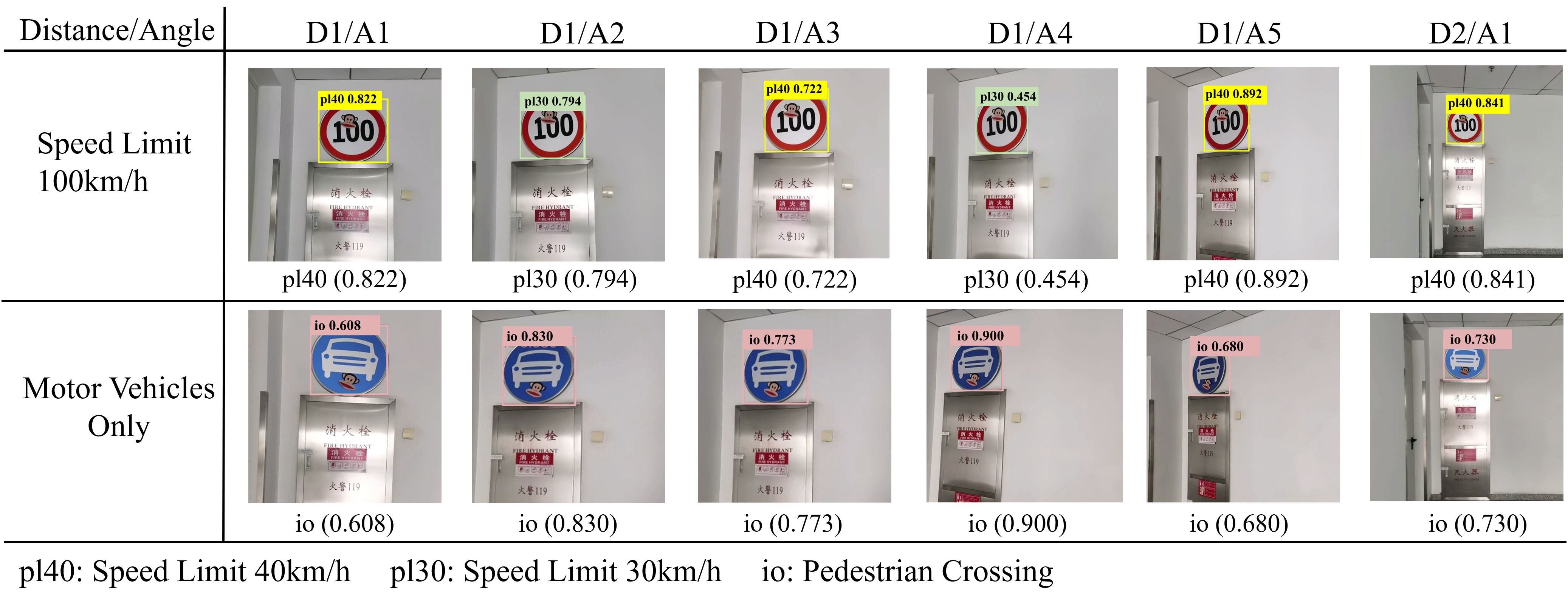}
\caption{ {{Examples of adversarial attacks for traffic sign recognition in different physical conditions including angles and distances. Roughly speaking, D1-D2 denote 1m, 3m, respectively, and A1-A5 denote $0^{\circ},5^{\circ},10^{\circ},15^{\circ},20^{\circ}$, respectively. The black text at the bottom of the image denotes the predicted class and its associated probability (shown in brackets)}}}
\label{fig:pp3}
\end{figure*}

\subsection{Traffic sign recognition}
To test the performance, we use the Tsinghua-Tencent 100K (TT100K) dataset \cite{zhu2016traffic} to conduct adversarial attacks. It provides 100,000 images containing 30,000 Chinese traffic-sign instances, which cover large variations in illuminance and weather conditions. We use  {the} YOLO \cite{redmon2016you} object detector as the threat model. We train YOLO using the training dataset in TT100K, and then test it  on 1000 images randomly selected from the testing dataset in TT100K. The final YOLO achieves 0.82 mAP. 

We firstly give the attacking performance in the digital world.  Some results are illustrated in Figure \ref{fig:multiple1} (classification attack) and Figure \ref{fig:multiple2} (disappearance attack).  In Figure \ref{fig:multiple1}, YOLO detector can successfully detect and classify the traffic sign for the original images, but mis-classify them for the adversarial images. In Figure \ref{fig:multiple2}, YOLO detector even fails to detect the traffic sign region. These results show that our adversarial stickers are effective for  {the} object detection task. In these adversarial examples, our stickers are small and a majority of them {do not} shelter from the foreground region in the traffic sign.  The quantitative attack performance on TT100K for traffic sign recognition using  {an} adversarial sticker is given in Table \ref{tab:Performance34},  where we can see that the mAP is significantly dropped from the original 0.82 to the current 0.23 after performing adversarial attacks. The drop error reaches 0.59 (the corresponding fooling rate is 85\%). This result shows the effectiveness of the adversarial sticker against  {the} traffic sign recognition task. The average query time is 435, which is also efficient. 

We also give the attacking performance in the physical world. For that, we firstly paste the sticker on the real traffic sign using the pre-computed position, and then use  {the} camera to capture the image  {at} different distances and {angles}. The captured image is finally fed to  {the} YOLO detector to predict the result. {Figure \ref{fig:pp3} shows the attack performance of the two signs at different angles and distances. For the traffic sign ``Speed Limit 100", the top row fools the object detector to predict it as ``Speed Limit 40 or 30'', and the object detector in the second row predicts ``Motor Vehicles Only'' as ``Pedestrian Crossing'', which verifies the effectiveness of adversarial stickers in the physical world.}

\section{Conclusion}
In this paper, we proposed the Meaningful Adversarial Sticker, a physically feasible and stealthy attack method for black-box attacks in the physical world. We conducted attacks based on the real stickers in our life by changing their pasting positions, rotation angles, and other parameters. To solve for the parameters efficiently, we designed RHDE algorithm, which adopted the offspring's generation strategy based on the aggregation of effective solutions and the adaptive adjustment strategy of the  evaluation criteria. Extensive experiments in the digital and  physical world on the face recognition, image retrieval, and traffic sign recognition demonstrated the effectiveness of our method. In the case that the model information is unknown, computer vision systems can also be successfully misled in a concealed way, which reveals the potential safety hazard.

\section*{Acknowledgments}
This work was supported by scientific research projects from HUAWEI Technologies Co. Ltd, and National Natural Science Foundation of China (No.62076018). We also thank anonymous reviewers for their valuable suggestions.

\ifCLASSOPTIONcaptionsoff
  \newpage
\fi

\bibliographystyle{IEEEtran}
\bibliography{arxiv}

\begin{IEEEbiography}[{\includegraphics[width=1in,height=1.25in,clip,keepaspectratio]{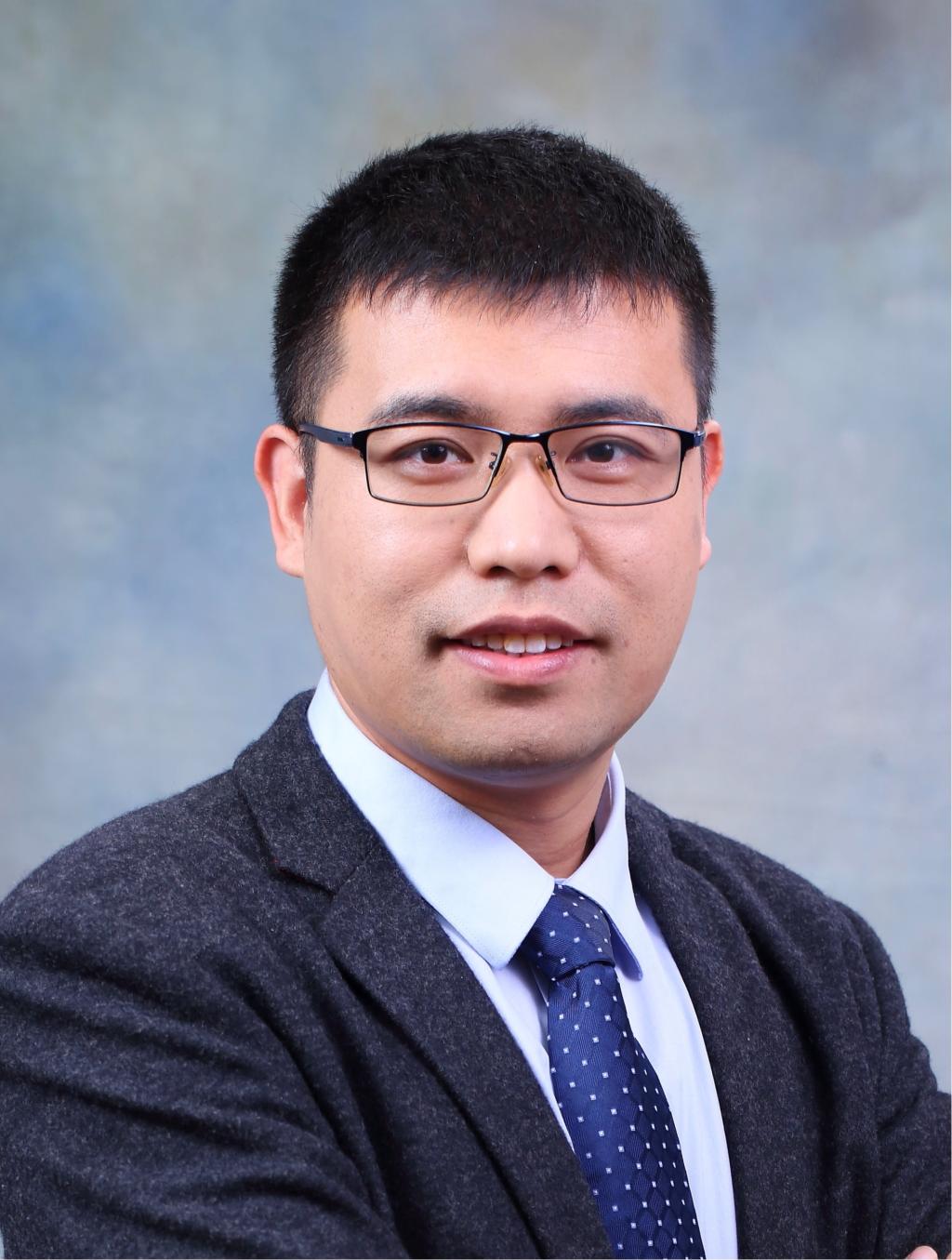}}]{Xingxing Wei}
received his Ph.D degree in computer science from Tianjin University, and B.S. degree in Automation from Beihang University, China. He is now an Associate Professor in Beihang University (BUAA). His research interests include computer vision, adversarial machine learning and its applications to multimedia content analysis. He is the author of referred journals and conferences in IEEE TPAMI, TMM, TCYB, TGRS, IJCV, PR, CVIU,  CVPR, ICCV, ECCV, ACMMM, AAAI, IJCAI etc.
\end{IEEEbiography}
\vspace{-1cm}
\begin{IEEEbiography}[{\includegraphics[width=1in,height=1.25in,clip,keepaspectratio]{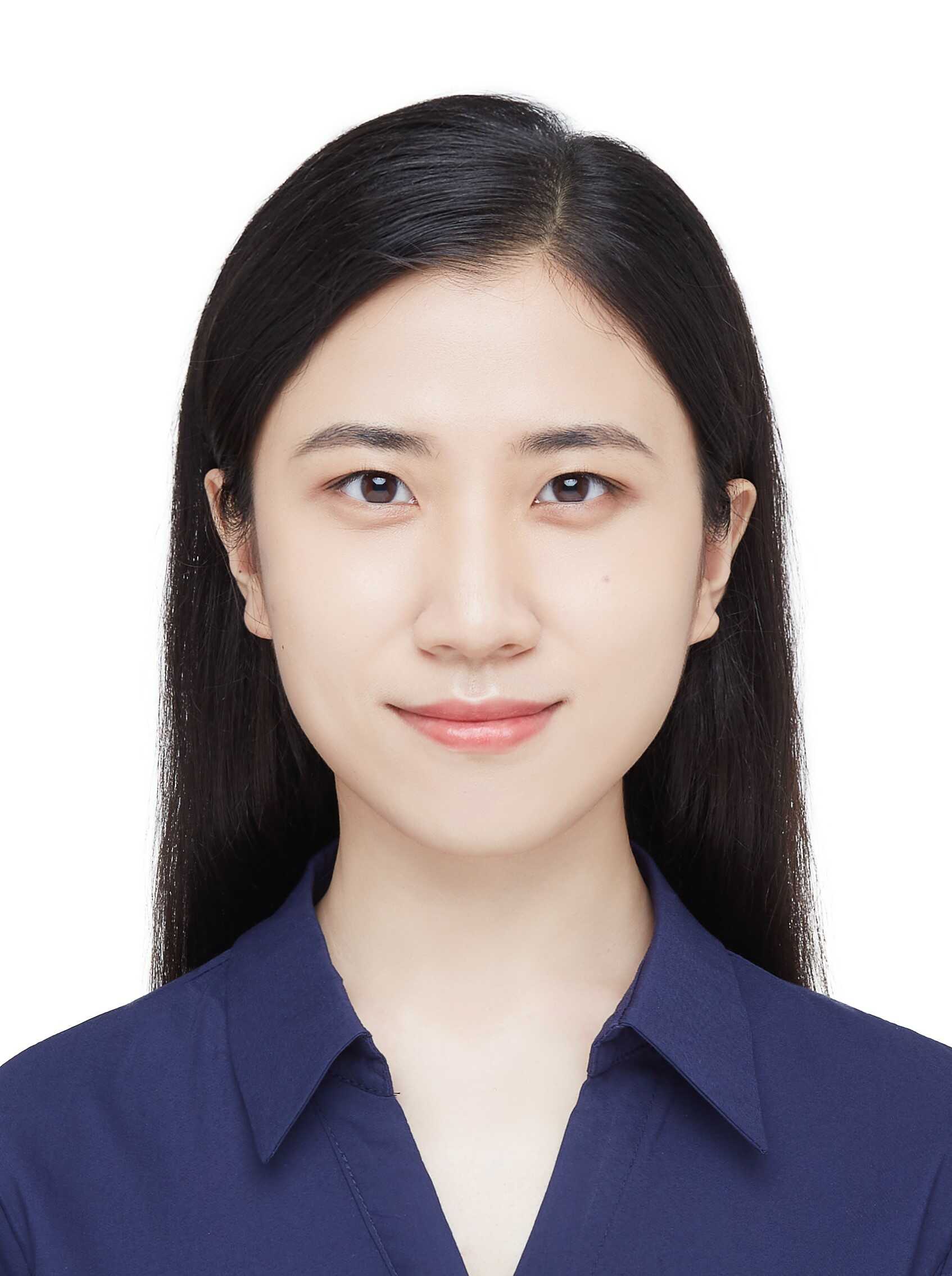}}]{Ying Guo} is now a Master student at School of Computer Science and Engineering, Beihang University (BUAA).
Her research interests include deep learning and adversarial robustness in machine learning.
\end{IEEEbiography}
\vspace{-1cm}
\begin{IEEEbiography}[{\includegraphics[width=1in,height=1.25in,clip,keepaspectratio]{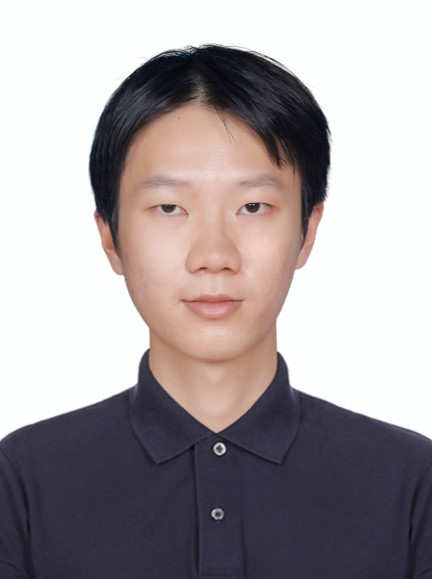}}]{Jie Yu} is now a Master student at School of Computer Science and Engineering, Beihang University (BUAA).
His research interests include deep learning, compter vision and adversarial robustness.
\end{IEEEbiography}

\end{document}